\documentclass[letterpaper, 10 pt, journal,twoside]{IEEEtran}  

\IEEEoverridecommandlockouts                              

\usepackage{graphics} 
\usepackage{epsfig} 
\usepackage{mathptmx} 
\usepackage{times} 
\usepackage{amsmath} 
\usepackage{amssymb}  
\usepackage{kotex}
\usepackage{color}
\usepackage{multirow}
\usepackage{subfigure}
\usepackage{cases}
\usepackage{algorithm}
\usepackage{algorithmic}

\usepackage[numbers]{natbib}
\bibpunct{[}{]}{,}{n}{}{;}
\usepackage[noadjust]{cite}  

\graphicspath{{./images/}}
\DeclareGraphicsExtensions{.pdf,.png,.jpg}

\definecolor{orange}{rgb}{1,0.2,0}
\definecolor{OliveGreen}{rgb}{0,0.6,0}

\newcommand{\yh}[1]{{ {#1}}}  

\newtheorem{designassumption}{\small \textbf{Design Assumption}}

\title{
	Non-linear Hysteresis Compensation of \\a Tendon-sheath-driven Robotic Manipulator\\ using Motor Current 
}

\author{Dong-Ho Lee$^{1}$, Young-Ho Kim$^{2*}$, Jarrod Collins$^{2}$, Ankur Kapoor$^{2}$, Dong-Soo Kwon$^{1}$, and Tommaso Mansi$^{2}$  
	\thanks{Manuscript received: October, 15, 2020; Revised December, 22, 2020; Accepted January, 20, 2021.}
	\thanks{This paper was recommended for publication by Editor Pietro Valdastri upon evaluation of the Associate Editor and Reviewers' comments.} 
	\thanks{$^{1}$Korea Advanced Institute of Science and Technology, Daejeon, South Korea
		{\tt\small vanquisher90@gmail.com, kwonds@kaist.ac.kr}}%
	\thanks{$^{2}$Siemens Healthineers, Digital Technology \& Innovation, Princeton, NJ, USA
		{\tt\small\{young-ho.kim,jarrod.collins,ankur.kapoor, tommaso.mansi\}@siemens-healthineers.com}}%
	\thanks{$^*$Corresponding author: {young-ho.kim@siemens-healthineers.com}}
	\thanks{Digital Object Identifier (DOI): see top of this page.}
}

\begin{document}
	
	\maketitle
	
	\begin{abstract}
		Tendon-sheath-driven manipulators (TSM) are widely used in minimally invasive surgical systems due to their long, thin shape, flexibility, and compliance making them easily steerable in narrow or tortuous environments. 
		Many commercial TSM-based medical devices have non-linear phenomena resulting from their composition such as backlash hysteresis and dead zone, which lead to a considerable challenge for achieving precise control of the end effector pose. However, many recent works in the literature do not consider the combined effects and compensation of these phenomena, and less focus on practical ways to identify model parameters in realistic conditions.
		\yh{This paper proposes a simplified piecewise linear model to construct both backlash hysteresis and dead zone compensators together. Further, a practical method is introduced to identify model parameters using motor current from a robotic controller for the TSM. 
			Our proposed methods are validated with multiple Intra-cardiac Echocardiography (ICE) catheters, which are typical commercial example of TSM, by periodic and non-periodic motions. Our results show that the errors from backlash hysteresis and dead zone are considerably reduced and therefore the accuracy of robotic control is improved when applying the presented methods.}
		
		
	\end{abstract}

	
	\vspace{-5pt}
	\section{INTRODUCTION}
	
	\IEEEPARstart{T}{he} tendon-sheath mechanism (TSM) is a popular control method that has been applied in many therapeutic \citep{daoud1999ep, khoshnam2015robotics, khoshnam2017robotics, bai2012worldwide, khan2013first} and real-time diagnostic (\textit{e.g.,} endoscope \citep{ott11endoscopy,le16survey,dario03review,phee97locomotion,lee21endoscopy}, colonoscope \citep{chen06kinematics}, and Intra-cardiac Echocardiography \citep{loschak16ice,kim2020automatic}) manipulators to achieve steerability by providing a long, thin, flexible structure that is compliant with anatomy. These TSM-based steerable manipulators are favorable in narrow and tortuous conditions, which makes them well-situated in relation to the growing shift towards minimally invasive treatment. 
	
\begin{figure}[t]
		\begin{center}
			\vspace*{5pt}
			\includegraphics[scale= 0.35]{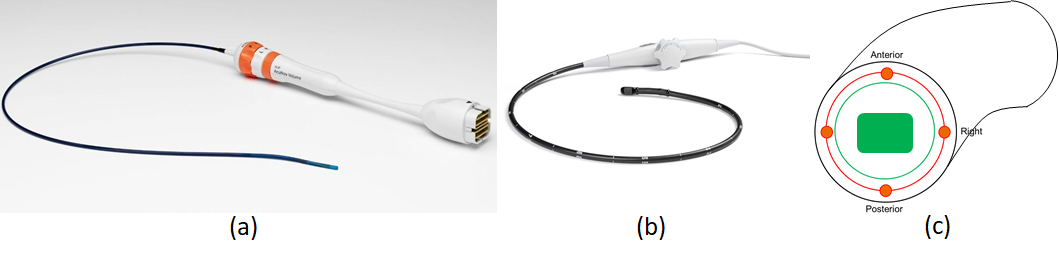}
			\caption{(a)-(b) Illustrative examples of 2-DOF TSM-based manipulators: ACUSON AcuNav
				Volume ICE catheter, ACUSON V5Ms TEE probe. Source: Siemens Healthineers. (c) 2-DOF TSM-based body cross-section: four thread sections (orange) combining with polymer cover (black) and ultrasound array (or tools) section (green) \label{fig:devices}}
			\vspace*{-20pt}
		\end{center}
	\end{figure}

	While the TSM-based manipulator has many advantages and wide adoption, the performance is still limited by non-linear frictional behaviors caused by: 1) backlash hysteresis due to friction forces between the sheath and tendons, 2) dead zone due to structural wire slack in the driving parts, and 3) plastic torsion due to the complex arrangement of threads and tubes about the {primary axis} of the device. The common structures of TSM is shown in Figure\,\ref{fig:devices}(c). These factors contribute to the degradation of control accuracy and limit the potential performance of robotic controllers for off-the-shelf TSM-based devices ({\em e.g.,}Intra-cardiac Echocardiography (ICE), Transesophageal Echocardiography (TEE) shown in Figure\,\ref{fig:devices}(a)(b)). 
	
	Accordingly, precise prediction of the tool tip pose for a tendon-sheath-driven robotic manipulator is challenged by these non-linear properties. An external robotic control system would therefore need to calibrate these effects before accurate manipulation can be achieved. Whether robotic control is being considered for disposable or reusable TSM manipulators, calibration would be required before each use. 
	However, there are practical limitations (\textit{i.e.} sterilization, cost, and size) which restrict adding traditional sensors to the tool tip to provide the necessary feedback for closed-loop control. As such, the control strategy for such a robotic system is open loop with no spatial feedback. However, much research has focused on modeling TSM itself without consideration of practical constraints and needs.

	
	
	
	\begin{figure}[t]
		\begin{center}
			\hspace{-5pt}
			\includegraphics[scale= 0.265]{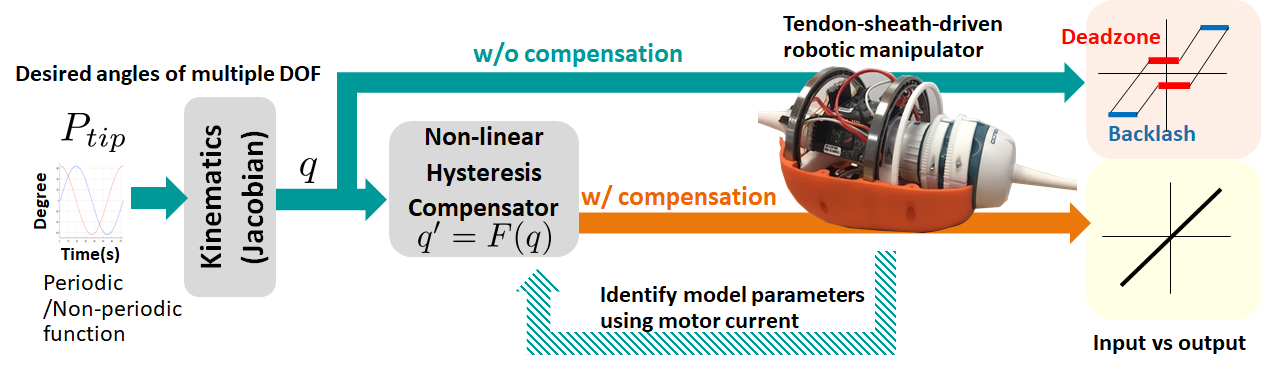}
			\vspace{-10pt}
			\caption{A representative diagram and scenario: Given desired pose at a given time, our goal is to compensate the configuration states, finding a compensated motions that minimizes errors. The compensated configuration ${\bf q'}$ gives input equal to output while ${\bf q}$ without compensation generates dead zone and backlash. A novel ideas are that 1) a simplified hysteresis model is proposed for both dead zone and backlash. 2) the model parameters are mainly identified using motor current based on motion behaviors. \label{fig:intro}}
			\vspace{-20pt}
		\end{center}
	\end{figure}

	\yh{This paper introduces a new method to model non-linear hysteresis, and a practical method for calibration of application in robotic control for TSM manipulators. More specifically, (1) a simplified piecewise linear model is proposed to compensate non-linear hysteresis of both backlash and dead zone together, and (2) in response to limitations in current practical settings, the relationship between non-linear hysteresis and motor current is experimentally investigated. Then, a practical parameter identification method is proposed, which associates motor current to particular motion behavior. Finally, the proposed methods are evaluated by periodic/non-periodic input motions for multiple catheters.}

	\section{Related Work}\label{sec:related}

	This section focuses on relating recent technical literature about TSM hysteresis modeling. Many researchers have addressed the non-linear characteristics of TSM with various analytical models ({\em e.g.,} friction, hysteresis), {as well as} image-based and data-driven approaches.
	
	Many studies have proposed a static model using coloumb friction\,\citep{kaneko91tendon,kaneko92tendon,chiang09tendon,fuxiang08tendon,chen12tendon,palli06tendon,palli06optimal} to overcome backlash hysteresis. Tension propagation is represented with a friction coefficient and the radius curvature (the shape) of the sheath. However, there exist limitations {when considering} dynamic motions.
	
	
	Various mathematical models including differential equations such as Bouc-Wen model and Prandtl-Ishlinskii model have also been proposed to reflect the dynamic characteristics \citep{radu09identification,hassani13piezo,do14hysteresis,do15adaptive}. However, there exist many hyperparameters, and the parameter identification is complicated. Mostly, additional sensors are required with controlled environments. In addition, they focus on varied shapes of backlash hysteresis rather than considering dead zone. The enhanced Bouc-Wen model considers the dead zone like shape (called as pinching\,\citep{Ismail2009ACME}), however their main module ({\em i.e.} energy function) is for structural engineering application ({\em{e.g.,}} vibration, stress modeling), which is not relevant to continuously manipulating system. 
	
	Image-based method is also proposed\,\citep{paolo17adaptive,reiter14appearance,baek20hysteresis}. This method is more robust than the previous methods in which the model's performance is affected when the shape of the sheath changes. Pose of the bending section is estimated through the obtained image, and feedback compensation is performed using the difference between the predicted bending angle and the input bending angle. However, the performance may change depending on the image quality or the presence of obstacles. 
	
	In order to overcome the dead zone, a data-driven method has been proposed\,\citep{yoon13error,zhang14flexible}. Data is obtained by sweeping the bending section up/down and left/right before use, and the motion was compensated by mapping the input and output data. However, this method also always required an additional sensor before use. 
	
	Although various attempts have been made, additional sensors such as load cells, vision sensors, and encoders were required, and it is {difficult} to attach additional sensors in a clinical environment. Also, only one degree of freedom is considered, and no studies have considered both backlash hysteresis and dead zone together.

	\section{Materials and Methods}

	\subsection{Tendon-sheath-driven robotic manipulator}
	
	An overview of the robotic system is illustrated in Figure\,\ref{fig:intro}. This is a typical open-loop control diagram for tendon-sheath-driven robotic  manipulators.
	Herein, the paper focuses on how to model and compensate the non-linear behaviors. Therefore, fundamental forward and inverse kinematics are not handled in this paper. The detailed kinematics models can be found in \citep{loschak16ice,kim2020automatic,kai08continuum}.
	
	Our motorized system can manipulate multiple degree-of-freedom (DOF) tendon-sheath-driven devices, which is briefly reviewed. The robot has four DOFs; two DOFs for steering the tip in two planes (anterior-posterior knob angle $\phi_1$ and right-left knob angle $\phi_2$) using two knobs on the handle, and other two DOFs for bulk rotation and translation along the major axis of the catheter body. Since rotation and translation do not contribute to the hysteresis phenomenon, the paper focuses on the two steering knobs. The robot's configuration state is defined as ${\bf q}=(\phi_1,\phi_2)$ in $\mathbb{R}^2$.
	
	Figure\,\ref{fig:intro} also shows an exemplary scenario. First, the desired pose of the tip $P_{tip}$ is given. Second, the desired robotic configuration state ${\bf q}$ is computed from the inverse kinematics model. Next, our compensator $F$ is applied to compute the compensated motor configuration ${\bf q'}$, which is directly applied to the motors. Then, the input versus the real output curve ideally shows a diagonal line for ${\bf q'}$ (\textit{i.e. }when properly compensated) while ${\bf q}$ (\textit{i.e.} without compensation) might show a hysteresis curve including deadzone and backlash. 
	
	Since external sensors ({\em e.g.,} load cell, electromagnetic (EM) tracker) are not considered in the field, it is evident the only input that we can use is motor current relative to enacted motions. Thus, the following section (Section\,\ref{sec:motorcurrent}) focuses on analysis of a relationship between the motor current and hysteresis curve.

	\begin{figure}[t]
		\centering
		\hspace{-10pt}
		\includegraphics[width= 9.3cm]{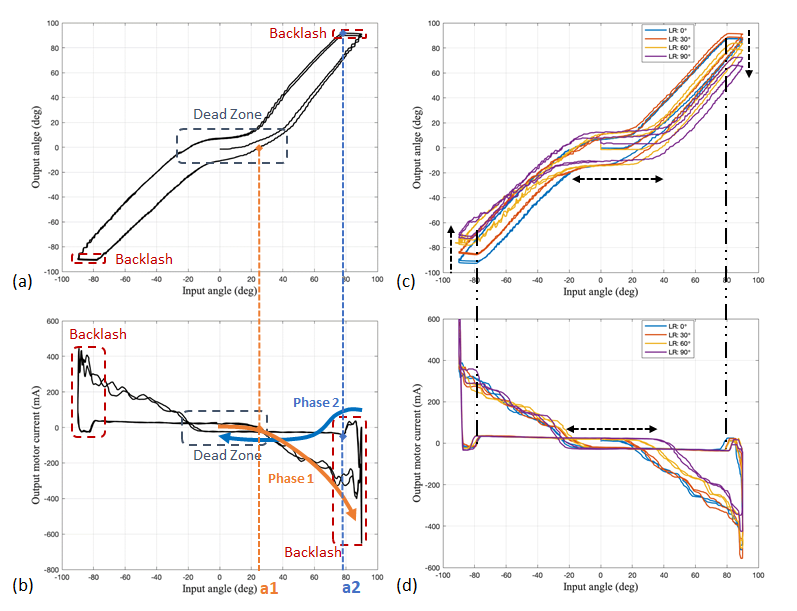}
		\caption{{\bf Left Column: } One analytical data represents dead zone and backlash hysteresis: (a) the desired robot state $\phi_1$ versus the real output angle  $\phi_1^{EM}$ (b) the desired robot state $\phi_1$ versus the motor current $c_1$. Two points ${\bf a1}$ (orange dotted) and ${\bf a2}$ (blue dotted) are marked to explain the dead zone and backlash. {\bf Right Column: } Sweeping motion for $\phi_1$ when $\phi_2$ is varied positively (equal to $0, 30, 60, 90^\circ$); (c) and (d) shows hysteresis curves and motor currents, respectively.}
		\vspace{-20pt}
		\label{fig:merged_relationship}
	\end{figure}

	\subsection{Systematic analysis of motor current and hysteresis curve } \label{sec:motorcurrent}
	
	In order to find out the relationship between the non-linear behaviors and the motor current, {a systematic test is conducted with simple sweep motions in the form of a sinusoidal wave, which has been commonly used in other studies} \citep{palli06tendon, hassani13piezo, do14hysteresis, baek20hysteresis}. The sweeping angle range is ${\pm 90^{\circ}}$, and data for $\phi_1$ and $\phi_2$ are collected. Two cycles of sweeping motions are applied with 0.04 Hz, and the shape of the sheath is constrained to maintain straight.

	
	The following data with the sinusoidal wave motion for each knob is collected: (1)~the desired robot configuration input ($\phi_1,\phi_2$), (2)~the real output angle of the bending section using EM tracker (3D guidance, Northern Digital Inc.), ($\phi_1^{EM},\phi_2^{EM}$), and (3)~the motor current ($c_1, c_2$) acquired from motor drivers in real time. The proper filter is applied for all settings (3rd order Butterworth filter, cutoff frequency 20 hz). 
	\yh{Figure\,\ref{fig:merged_relationship} demonstrates one representative data to understand the relationship between the non-linear behaviors and motor current; Figure\,\ref{fig:merged_relationship}(a)(b) shows one analytical data representing the desired input angle versus the real output angle, and the desired input angle versus the real output, respectively. Figure\,\ref{fig:merged_relationship}(c)(d) shows overlaid analytical data in different conditions of other DOF.}
	
	
	{\noindent~{\bf Lesson~1} from~Figure\,\ref{fig:merged_relationship}(a)}: 
	An ideal TSM will show the desired input and the real output should be same during operation. However, multiple non-linear behaviors are observed in practice: (1)~{\it dead zone: } when the input angle is near zero, there is a dead zone that maintains a constant output value even if the input value increases (or decreases). (2)~{\it backlash hysteresis: } when the direction of motion is changed, there is delay in the real output angle rather than immediately increasing (or decreasing) the angle. 
	
	
	
	
	
	{\noindent~{\bf Lesson~2} from~Figure\,\ref{fig:merged_relationship}(b)}: To explain the relationship of motor current and non-linear behavior, the sweeping motion is classified into two phases; {\it Phase~1 (dead zone):~} This is an interval from 0$^{\circ}$ to 90$^{\circ}$ of the input angle.
	As the input is increased from 0$^{\circ}$ to a certain input angle, {\it a1} shown in Figure\,\ref{fig:merged_relationship}(b), \yh{there exists a smooth flat signal shape, which means the motor current magnitude remains a constant value until {\it a1}. Accordingly, the output angle is not changed according to the input changes in {\it Phase~1}.
		However, when exceeding {\it a1} , As the input angle is gradually increased, the output angle is also increased and the motor current is significantly increased.}
	The same phenomenon is observed when moving in the opposite direction.
	\yh{Thus, based on results of experimental validation, it is evident that the motor current can be used to detect the width of the dead zone.}
	{\it  Phase~2 (backlash):~} This is an interval from 90$^{\circ}$ to 0$^{\circ}$. The moment when the direction of the desired input changes shows an interesting phenomenon such that the output angle is maintained for a while before {gradually decreasing}. Looking at the behavior of the motor current at this moment, the motor current direction changes sharply in the opposite direction and remains a constant value close to zero after making a small peak.
	\yh{The reason for the peak is happened when the applied input angle direction is changed from increments to decrements. At the moment, the wire is from being pulled to being released, then the tension does not immediately switch in the opposite direction, but goes through a transitional phase\,\citep{chiang09tendon,chen12tendon,palli06tendon}, which generates a small motor current peak until the wire slack is disappeared. Then, the motor current stays a certain current level as tension is balanced because the bending section is a continuum, thus it has the property to be straighten itself without the help of the motor.
		The experimental results indicate that {\it a2} at which the current begins to become constant is associated with the end of the backlash interval. Thus, it is apparent that the motor current measures can be used to detect the backlash interval.}

	{\noindent~{\bf Lesson~3} from~Figure\,\ref{fig:merged_relationship}(c)(d)}: \yh{Additional experimental analysis is used to understand the dead zone and backlash changes with regard to other DOF changes.
		One knob $\phi_1$ is swept, while another knob $\phi_2$ is fixed at a constant value, but this keeps changed from $0^{\circ}$ to $ 90^{\circ}$ with 30$^{\circ}$ intervals.}
	Figure\,\ref{fig:merged_relationship}(c) shows the dead zone is shifted as the fixed value of $\phi_2$ is increased. The same phenomenon is occurred with the motor current in Figure\,\ref{fig:merged_relationship}(d). 
	In the case of the backlash hysteresis, there is no significant change in either graph. However, it is observed that the backlash hysteresis ends and the motor current remains constant after the small peak is observed.
	Furthermore, in Figure\,\ref{fig:merged_relationship}(c), the slopes of the output are similar except for the dead zone and backlash period, and also the output angles in the dead zone are similar. 
	The opposite direction, \textit{i.e. }when $\phi_2$ is decreased from 0$^{\circ}$ to -90$^{\circ}$, is not shown here, but the result is symmetric, which is biased to the negative side.

	\subsection{Modeling non-linear hysteresis using piecewise linear approximation}
	
\begin{figure}[t]
		\centering
		\hspace{-10pt}
		\includegraphics[scale=0.65]{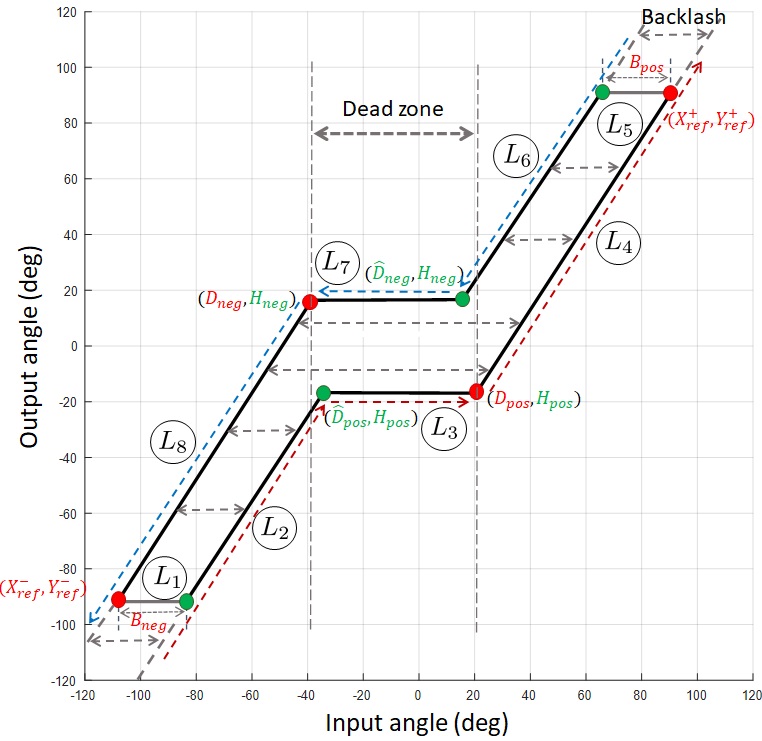}
		\caption{\yh{Our proposed piecewise linear model $L$: eight equations represent non-linear hysteresis phenomena. (1) Four main parameters in red colored: Identification of the size of the backlash hysteresis ($B_{pos}$, $B_{neg}$), and the range of the dead zone ($D_{pos}$, $D_{neg}$) are acquired from the motor current data.
				(2) Associated parameters: $\hat{D}_{pos}$, $\hat{D}_{neg}$, $H_{pos}$, and $H_{neg}$ are computed based on known main parameters.
				(3) Working flows: When the velocity $\dot{x}_t$ is positive (red arrows), the output moves on lines $L_{2}$, $L_{3}$, and $L_{4}$, and when $\dot{x}_t$ is negative (blue arrows), the output moves on lines $L_{6}$, $L_{7}$, and $L_{8}$. When the direction of movement changes ($\dot{x}_{t} \cdot \dot{x}_{t-1} < 0)$ (bidirectional gray arrows), $X^+_{ref}$ (or $X^-_{ref}$ is updated based on $x_t$ and the output moves on lines $L_{1}$ and $L_{5}$. 
		}}\vspace{-15pt}
		\label{fig:approximated_simple_model}
	\end{figure}

	\yh{As discussed in Section\,\ref{sec:motorcurrent}, a highly non-linear object curve is related to the desired input and motor current. To simplify both the modeling and parameter identification problems, this paper proposes a piecewise linear approximation to represent the non-linear hysteresis phenomenon and parameter identification methods using motor current. The proposed model consists of a total of eight linear equations, half of which are when the velocity is positive and others are when the velocity is negative. 
		To define a finite collection of linear functions, 
		let $x_t\in {\bf q}$ and $\dot{x}_t\in {\bf \dot{q}}$ represent the one-DOF input state and velocity at time $t$, respectively, where $t$ is a temporal index. Four main parameters that can be acquired from motor current are defined; $D_{pos}$ and $D_{neg}$ represent the dead zone range for positive and negative directions. $B_{pos}$ and $B_{neg}$ represent the size of the backlash hysteresis for each direction. Six associated parameters are defined; $H_{pos}$ and $H_{neg}$ denote the height of the dead zone for each direction. $\hat{D}_{pos}$ and $\hat{D}_{neg}$ represent the opposite side of the dead zone $D_{pos}$ and $D_{neg}$, respectively. Additionally, $\omega$ denotes the slope of the lines, and let $(X^+_{ref}, Y^+_{ref})$ and $(X^-_{ref}, Y^-_{ref})$ denote reference points for each direction, which is the input and the real output state. 
		All parameters of our proposed method are shown in Figure\,\ref{fig:approximated_simple_model}}.
	
	
	
	Our philosophy is to identify parameters of proposed method benefit from motor current. Thus, three design assumptions are made:
	\begin{designassumption}
		The initial $X^{\pm}_{ref}$ and $Y^{\pm}_{ref}$ are given when one DOF is swept while the other DOF is fixed to 0$^{\circ}$.
	\end{designassumption}
	\noindent \hspace*{10pt}{\it Rationale:}~ This is a reasonable assumption since the other DOF stays at 0 degree, so the interference from the other DOF can be minimized while one DOF is being swept. The commercial catheter is usually calibrated for $\phi_1$ and $\phi_2$ with regard to real output for reference axis like $\pm 90^\circ$.
	In this manner, it becomes fairly trivial to acquire a reference point in specific condition ({\em i.e.,} ($\phi_1$, $\phi_2$)=\{(0,90), (90,0), (-90,0)...\}) by visual inspection.
	
	\begin{designassumption}
		The slope of the lines, $\omega$ of the target device is constant.
	\end{designassumption}
	\begin{designassumption}
		The height of the dead zone, $H_{pos}$ and $H_{neg}$ are constant for the target device.
	\end{designassumption}
	\noindent \hspace*{10pt}{\it Rationale:}~ As our target device is a commercial product, so its physical properties are optimized as similar. $\omega$ and $H$ can be brought from mechanical data sheets or can be measured one time for each product.
	
	\yh{Accordingly, our proposed hysteresis model $L(x_t,\dot{x_t},\dot{x}_{t-1})$ is determined by eight linear equations with four known parameters associated with given three design decisions.
		Figure\,\ref{fig:approximated_simple_model} shows how eight equations are divided and switched over the region. To construct linear functions, $\hat{D}_{pos}$ and $\hat{D}_{neg}$ are computed in Equation\,\eqref{eq:D_hats} given $D_{pos}$, $D_{neg}$, $B_{pos}$, $B_{neg}$, $H_{pos}$, $H_{neg}$, and $\omega$.}
	
	\yh{Then, all eight linear equations of Figure\,\ref{fig:approximated_simple_model} can be determined in Equation\,\eqref{eq:model_eqs}. The dead zone ($L_{3}$, $L_{7}$) intervals consist of a horizontal line at $H_{pos}$ and $H_{neg}$, respectively. The backlash hysteresis ($L_{1}$, $L_{5}$) intervals are a horizontal line, but updated with the current state $x_t$ when the direction is changed. The other intervals consist of a line with the slope of $\omega$.}

	{\vspace{-10pt}
		\begin{eqnarray}\label{eq:D_hats}
		\begin{aligned}
		\hat{D}_{pos} &= (H_{pos}-H_{neg})/\omega + D_{neg}+B_{neg} \\
		\hat{D}_{neg} &= (H_{neg}-H_{pos})/\omega + D_{pos}-B_{pos}
		\end{aligned}
		\end{eqnarray}\vspace{-15pt}
	}
	
	{
		\begin{equation}\label{eq:model_eqs}
		L(x_t,\dot{x}_t,\dot{x}_{t-1}) =
		\begin{cases}
		\text{$L_{1}$}:     \omega(-X_{ref}-D_{neg})+H_{neg} \\
		\text{$L_{2}$}: \omega(x-\hat{D}_{pos})+H_{pos}\\
		\text{$L_{3}$}: H_{pos}\\
		\text{$L_{4}$}: \omega(x-D_{pos})+H_{pos}\\
		\text{$L_{5}$}: \omega(X_{ref}-D_{pos})+H_{pos}\\
		\text{$L_{6}$}: \omega(x-\hat{D}_{neg})+H_{neg}\\
		\text{$L_{7}$}: H_{neg}\\
		\text{$L_{8}$}: \omega(x-D_{neg})+H_{neg}\\
		\end{cases}
		\end{equation}
	}

	\begin{figure}[b]
		\vspace{-10pt}
		\centering
		\includegraphics[width=9.1cm]{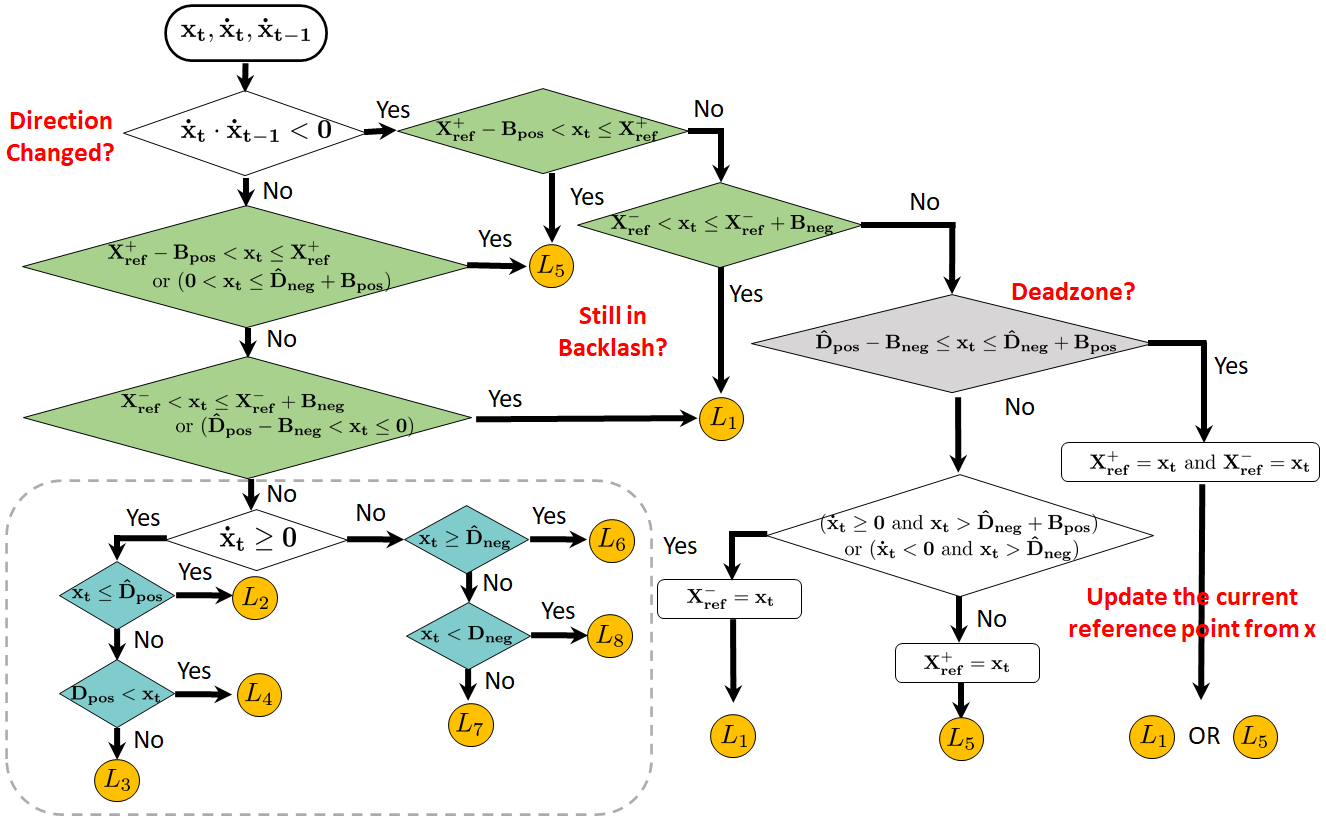}
		\caption{\yh{Model transition conditions for arbitrary inputs: The inputs $x_t$. $\dot{x}_t$, and $\dot{x}_{t-1}$ are used to decide which linear function is activated. The green colored conditions are to check $x_t$ in the backlash intervals. The gray colored condition is to check $x_t$ is in dead zone.
				The dotted rectangle (the blue colored conditions) is for a general operation when the direction is not changed.}}
		\label{fig:prediction}
	\end{figure}

	\yh{Figure\,\ref{fig:prediction} shows the detailed workflow diagrams, which use the current input state $x_t$, the current velocity $\dot{x}_{t}$, and the previous velocity $\dot{x}_{t-1}$ to determine which linear approximation is activated. First, if the direction is changed, there are two cases; 1) $x_t$ was already in the backlash, then $L_1$ (or $L_5$) can be easily picked. 2) $x_t$ was not in the backlash, then firstly. if $x_t$ is in dead zone, then $X^+_{ref}$ and $X^-_{ref}$ are updated, and selected one of $L_1$ and $L_5$, which is effective when $x_t$  is located inside deadzone. Otherwise, $X^+_{ref}$ (or $X^-_{ref}$) is updated according to $x_t$. Then, either $L_1$ or $L_5$ can be selected.}

	\yh{If the direction is not changed, there are mainly two conditions; 1) if $x_t$ was in the backlash, then either $L_5$ or $L_1$ can be easily found based on $x_t$. 2) Otherwise, depending on $\dot{x}_{t}$, either the positive direction ($L_2$, $L_3$, and $L_4$) or the negative direction ($L_6$, $L_7$, and $L_8$) is found, and the current state $x_{t}$ is used to determine which specific positive (or negative) linear function is called.}
	
	

	\yh{Then, our proposed hysteresis compensator $F$ can be obtained by using the inverse of the completed model $L$ defined as $F =L(x_t,\dot{x}_{t},\dot{x}_{t-1})^{-1}$, which can cover the whole workspace for multiple DOFs TSM manipulator and handle the arbitrary input signals. More specifically, when multiple DOFs are manipulated, one DOF model for $D_{pos}$, $D_{neg}$, $B_{pos}$, keeps updated depending on other DOF state $x_{t}$. The interpolated values of four parameters for multiple DOFs are explained in next Section\,\ref{sec:ex:calibration}.} Then, this compensation is used as a feed-forward control after receiving the desired input. Then, the desired input ${\bf q}$ can be compensated by ${\bf q}'$ as shown Figure\,\ref{fig:intro}.

	\subsection{Parameter Identification}\label{sec:calibration}
	
	
	In order to use the proposed model, the model parameters have to be identified for each target device. 
	Systematic motion behaviors for two knobs $(\phi_1,\phi_2)$ are designed to identify four main parameters; One DOF is swept over a sinusoidal wave with the range of ±90 degrees when the other DOF is fixed at constant value, which varies from $-90^{\circ}$ to $+90^{\circ}$ with $30^{\circ}$ intervals. Two  cycles  of  sweeping  motions  are  applied  with 0.04 Hz, and the shape of the sheath is constrained to maintain straight.
	
	{\bf \noindent Data collection:~}
	Three data are collected; (1)~the desired robot configuration input ${\bf q}$. (2)~the real output angle of the bending section using EM tracker, ${\bf q^{EM}}$. (3)~the motor current ($c_1, c_2$) acquired from motor drivers in real time. The proper filter for all settings (3rd order Butterworth filter, cutoff frequency 20 hz) are also applied.
	
	{\bf \noindent Calibration from experimental data:~}
	Then, the range of dead zone $D$ and the size of backlash hysteresis $B$ are extracted at each angle of the other DOF from motor current. The dead zone is obtained by selecting a moment when the period of constant current ends and it starts to increase or decrease, which is the point {\it a1} in Figure\, \ref{fig:merged_relationship}(b). To find this moment, the 'findchangepts' function in Matlab is used. This function returns the index at which the mean value changes the most in the data, so the moment when the slope changes sharply can be found. The size of the backlash hysteresis is obtained by selecting the moment at which the current starts to become constant after a small peak after changing the moving direction, which is shown as the point {\it a2} in Figure\, \ref{fig:merged_relationship}(b). To find this moment, the 'findpeak' function in Matlab is used. This function provides the local maxima (or minima) in the data.
	
	Other associated parameters are obtained: (1) The slope of the line ($\omega$) is obtained from the data collected for four ICE catheters. From the data, the mean value was 1.32 and the standard deviation was 0.12, this shows our design assumption-2 is valid. (2) The height of the dead zone ($H_{pos}, H_{neg}$) are obtained using the slope of the line and the reference point, which is passing the reference point and the dead zone ends, thus ($H_{pos}, H_{neg}$) can be computed. 
	(3) $\hat{D}_{pos}$,  $\hat{D}_{neg}$ are calculated based on Equation\,\eqref{eq:D_hats}.
	
\begin{figure}[t]
		\centering
		\includegraphics[scale=0.18]{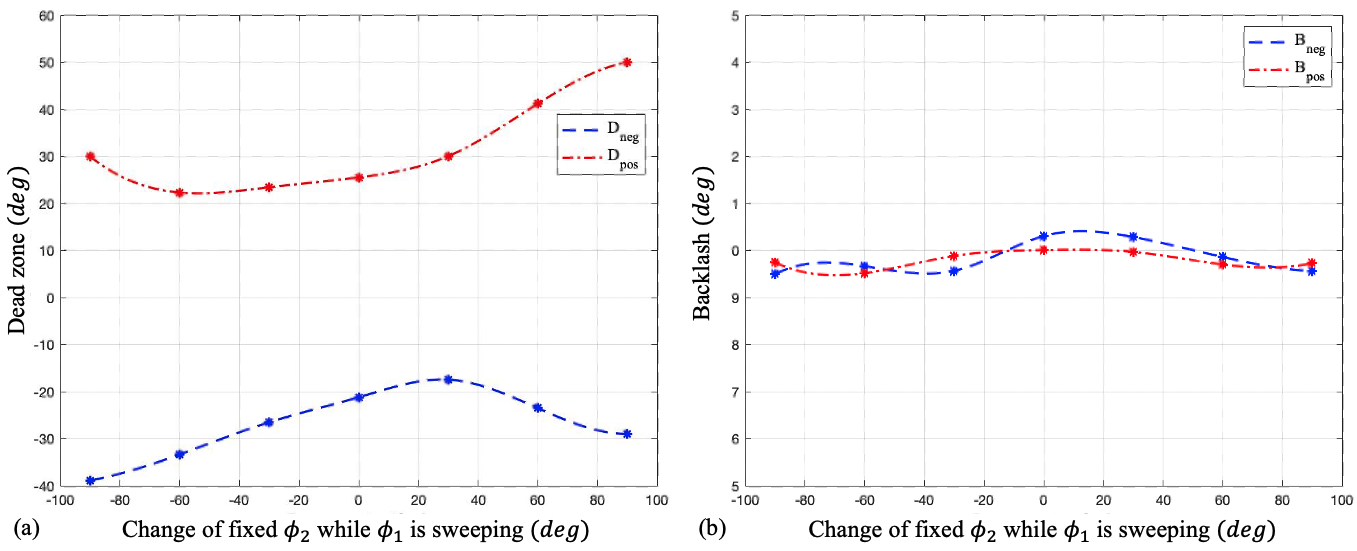}
		\caption{Anterior-posterior knob $\phi_1$ is sweeping over the whole workspace while left-right knob $\phi_2$ is fixed one condition ( $0, \pm30, \pm60, \pm90$ degrees). The dead zone ($D_{pos}$, $D_{neg}$) (a) and the backlash ($B_{pos}$, $B_{neg}$) (b) are interpolated using collected data (dotted points), thus four main parameters are estimated over a whole workspace.}
		\label{fig:regression}
	\end{figure}

	{\bf \noindent Interpolation of four main parameters:~}
	\yh{
		From discrete motion combination ($\pm 90^\circ$ at $30^\circ$ intervals), the dead zone $D_{pos}$, $D_{neg}$ and the backlash $B_{pos}$, $B_{neg}$ are interpolated using the collected data to have a continuous parameter space over the workspace.
		One real data is shown as an example in Figure\,\ref{fig:regression}. $D_{pos}$ shows a downward convex bell shape, and $D_{neg}$ shows an upward convex bell shape. $B_{pos}$ and $B_{neg}$ are almost similar values. 
		Accordingly, the model calibration is completed for the whole continuous workspace for multiple coupled DOFs, and other associated parameters are also updated.}

	\section{Experimental Result}
	\subsection{System setup}
	\yh{The ICE catheter shown in Figure\,\ref{fig:devices}(a) was selected as an example of TSM, attached to motorized system that can manipulate two DOFs for steering the tip of ICE catheter in two planes.
		An EM sensor (Model 800 sensor, 3D guidance, Northern Digital Inc.) was attached to the catheter tip to provide real-time tracking of position and orientation, which are used as a ground truth via inverse kinematics.}
	
	
	\subsection{Scenarios: controllers, input motions, conditions}\label{sec:scenarios}
	\yh{Three ICE catheters were tested as following scenarios with five times repeated trials.}
	\subsubsection{Two controllers}\label{sec:scenarios_controller}
	\yh{Two types of controllers were considered for comparison of the experimental results. (1) {\it ~No compensation:} This controller was designed to send $q$ directly to the motors without compensation. (2){\it ~With compensation:} This generated compensated $q'$ using our proposed compensator working as a feedforward compensation.}
	\subsubsection{Two input motions}\label{sec:scenarios_input}
	\yh{Two types of input motions were designed to validate our proposed methods. (1) {\it Periodic:} A periodic motion was designed as ${60^\circ}$ amplitude and 0.04~Hz. (2) {\it Non-periodic:} A non-periodic motion was designed as a combination of two sinusoidal signals with $30^\circ$ for both amplitudes and $0.02$~Hz and $0.02\sqrt{3}$~Hz in frequencies.
	}
	\subsubsection{Two operating conditions}\label{sec:scenarios_conditions}
	\yh{Two operating conditions were designed. (1) {\it One DOF operation} is designed to verify our proposed method in One DOF operation. Each input motion from Section\,\ref{sec:scenarios_input} was applied to each knob $\phi_1$ and $\phi_2$; One knob was operated while another DOF was fixed at a constant value (\textit{e.g.,} varied $0,~\pm30,~\pm60$). (2) {\it Two DOFs simultaneously operation} was designed as follows; one knob $\phi_1$ operation is same as {\it One DOF operation} while another knob $\phi_2$ is operated as the same input motions, but twice faster than the designed frequency.
	}
	
	
	
	\subsection{Validation metrics}\label{sec:ex:validation}
	\yh{To evaluate the proposed method, the magnitude of the peak-to-peak error (PTPE) is used, which is a difference between the highest and the lowest error. Moreover, the root mean squared error (RMSE) was used for performance evaluation, and repeatability is represented by mean and standard deviation.}
	
	
	\subsection{Model calibration}\label{sec:ex:calibration}
	\yh{As a first step, model parameters for each catheter were identified following the calibration procedure in Section\,\ref{sec:calibration}. The slope of the lines `${\it A}$' was assigned the average of the four ICE catheters, $1.32$. Further, $H_{pos}$, $H_{neg}$, $\hat{D}_{pos}$, and $\hat{D}_{neg}$ were calculated with the determined parameters.
	}

	\subsection{Result of the non-linear hysteresis compensation}
	
	Overall, three catheters were tested five times each for designed scenarios in Section\,\ref{sec:scenarios}., thus all errors are described with mean and standard deviation as $(\mu,\sigma)$ in the tables.
	
\begin{figure*}[t]
		\subfigcapskip = -4pt
		\begin{center}
			
			
			
			\hspace{-15pt}
			\subfigure[$\phi_1$ (or $\phi_2$) = $-60^\circ$]{\includegraphics[scale=0.10]{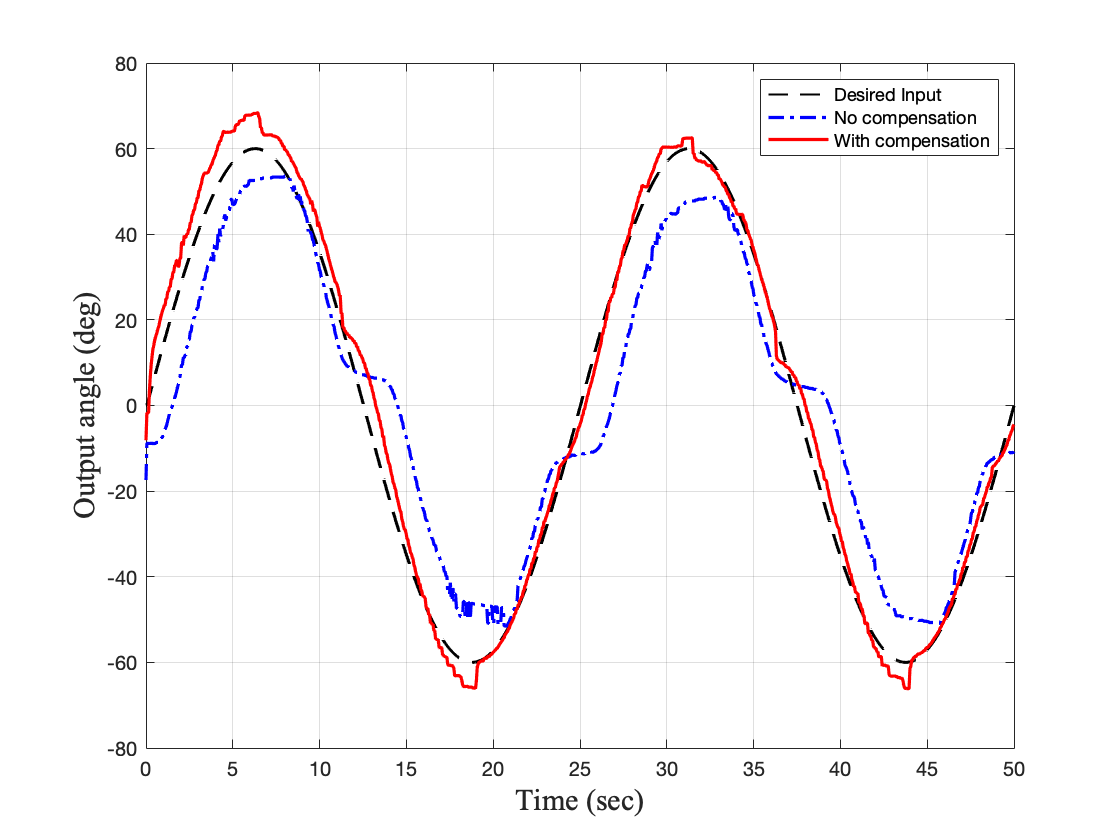}\label{fig:LR11}}
			\hspace{-15pt}
			\subfigure[$\phi_1$ (or $\phi_2$) = $-30^\circ$]{\includegraphics[scale=0.10]{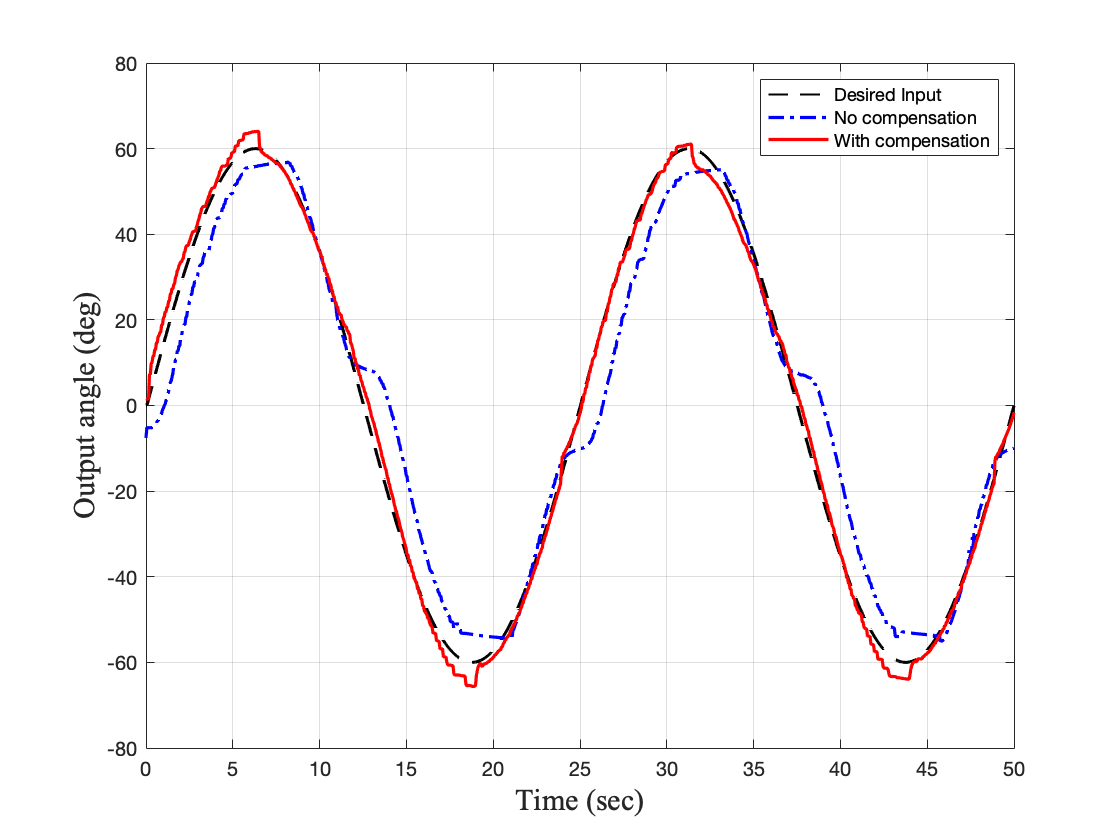}\label{fig:LR22}}
			\hspace{-15pt}
			\subfigure[$\phi_1$ (or $\phi_2$) = $0^\circ$]{\includegraphics[scale=0.10]{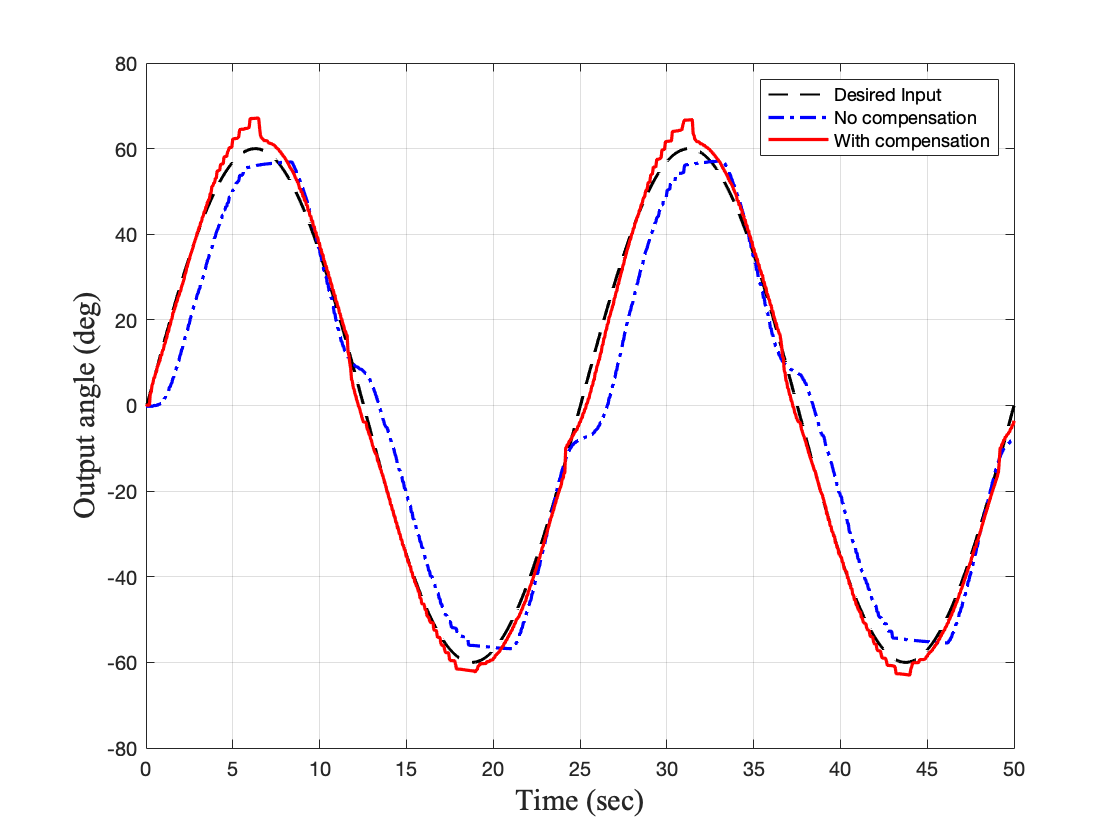}\label{fig:AP33}}
			\hspace{-15pt}
			\subfigure[$\phi_1$ (or $\phi_2$) = $30^\circ$]{\includegraphics[scale=0.10]{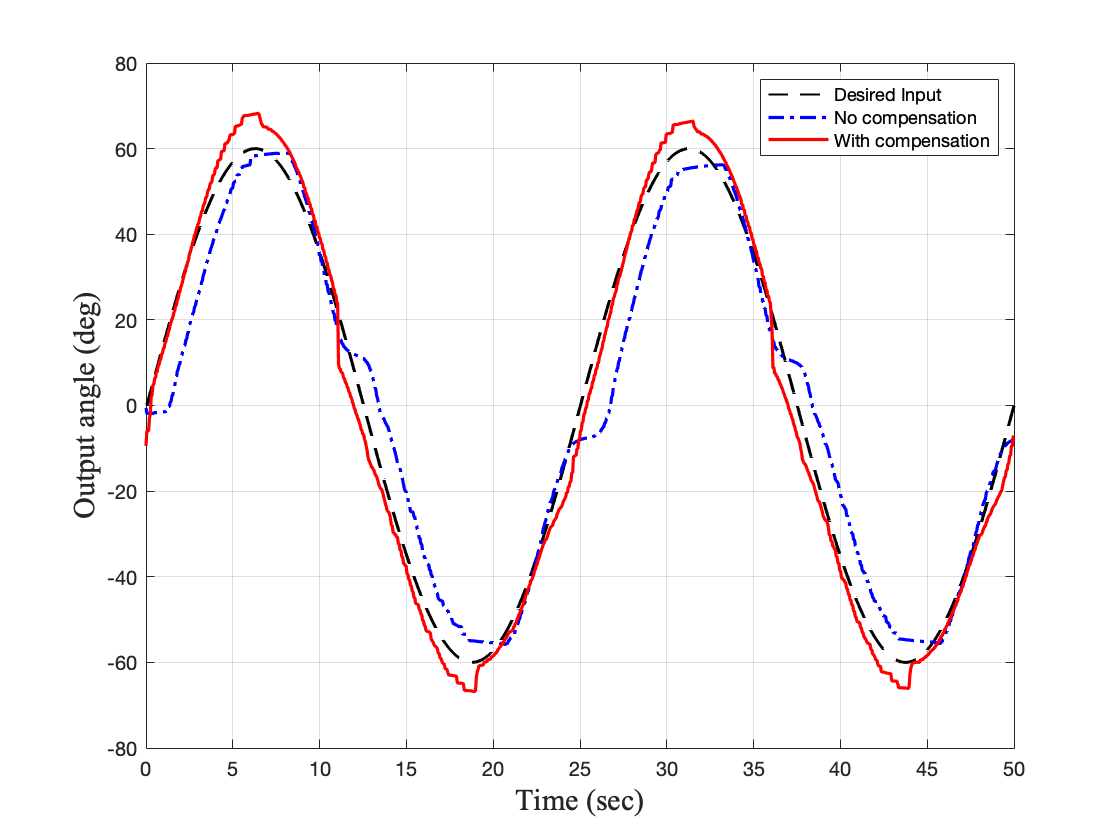}\label{fig:LR44}}
			\hspace{-15pt}
			\subfigure[$\phi_1$ (or $\phi_2$) = $60^\circ$]{\includegraphics[scale=0.10]{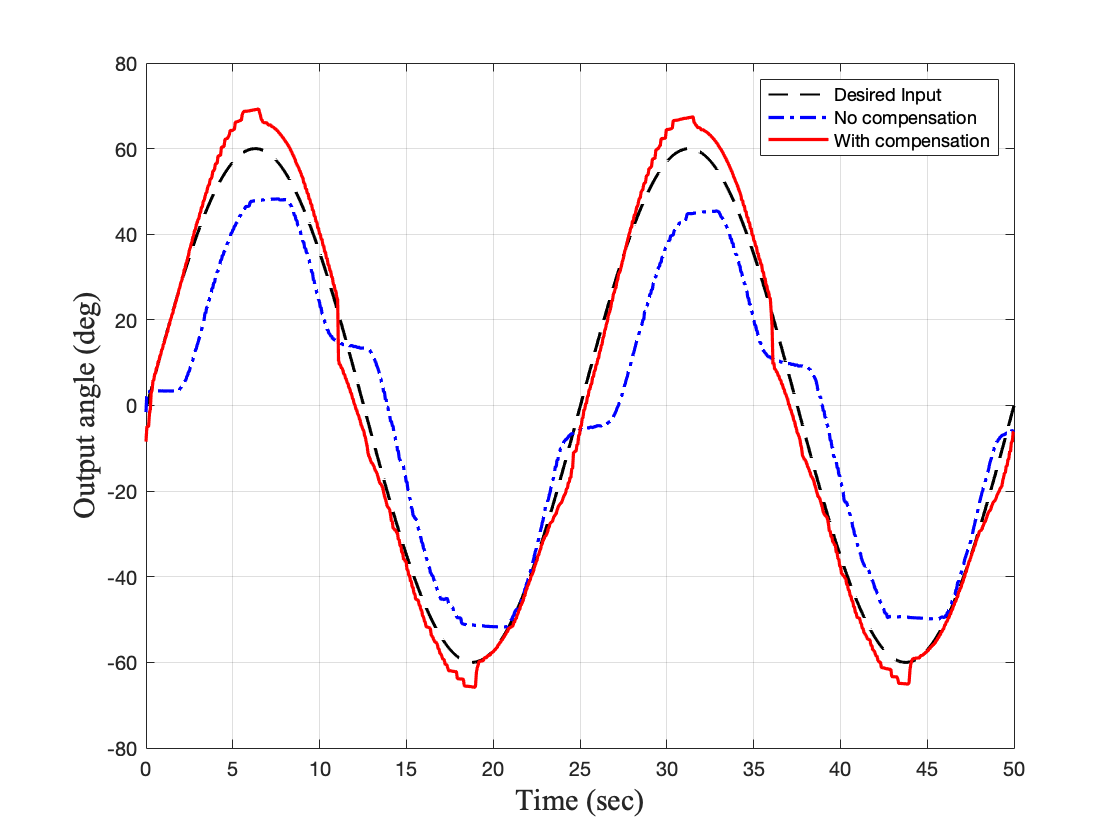}\label{fig:LR55}} \\
			
			\hspace{-15pt}
			\subfigure[$\phi_1$ (or $\phi_2$) = $-60^\circ$]{\includegraphics[scale=0.1]{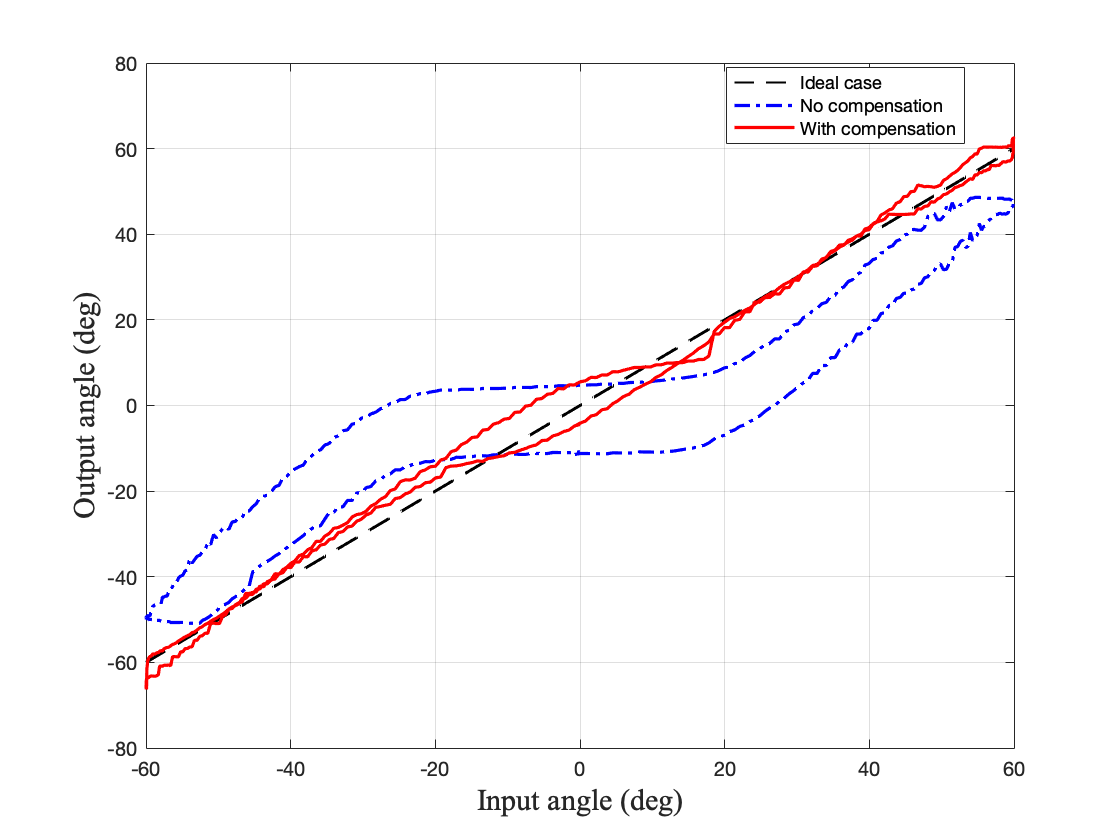}\label{fig:LR1}}
			\hspace{-15pt}
			\subfigure[$\phi_1$ (or $\phi_2$) = $-30^\circ$]{\includegraphics[scale=0.1]{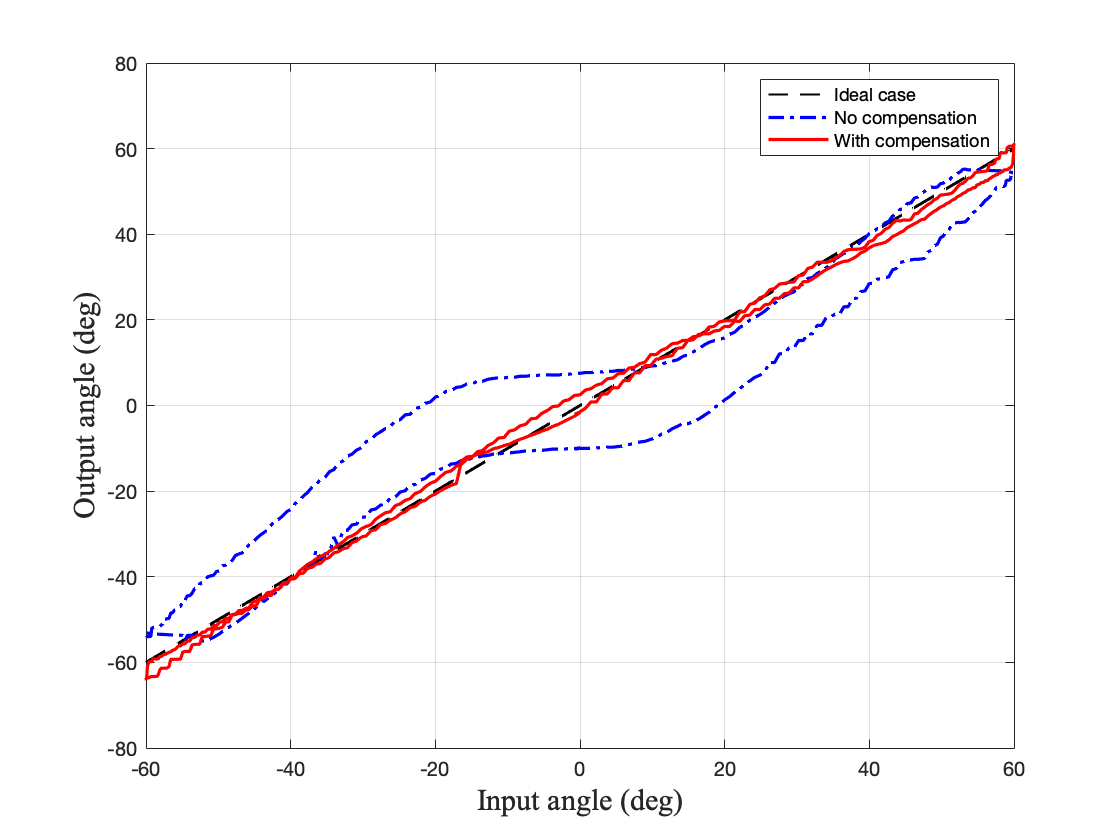}\label{fig:LR2}}
			\hspace{-15pt}
			\subfigure[$\phi_1$ (or $\phi_2$) = $0^\circ$]{\includegraphics[scale=0.1]{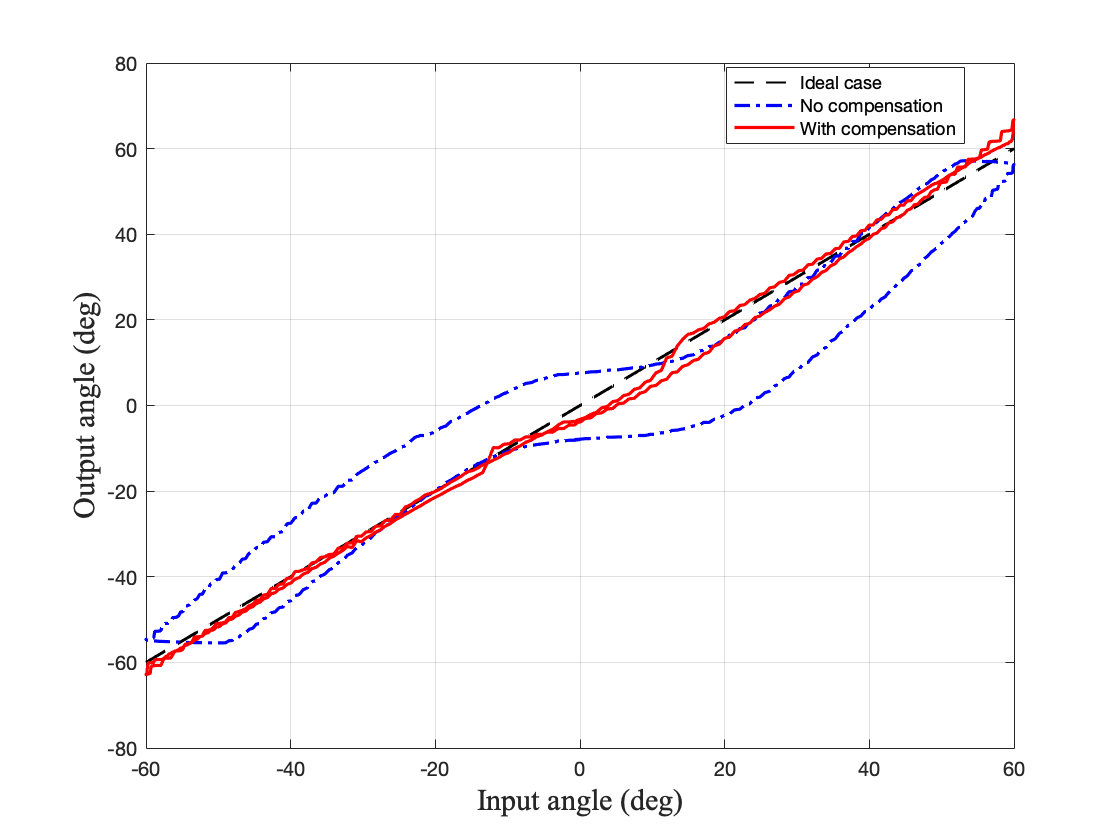}\label{fig:LR3}}\
			\hspace{-15pt}
			\subfigure[$\phi_1$ (or $\phi_2$) = $30^\circ$]{\includegraphics[scale=0.1]{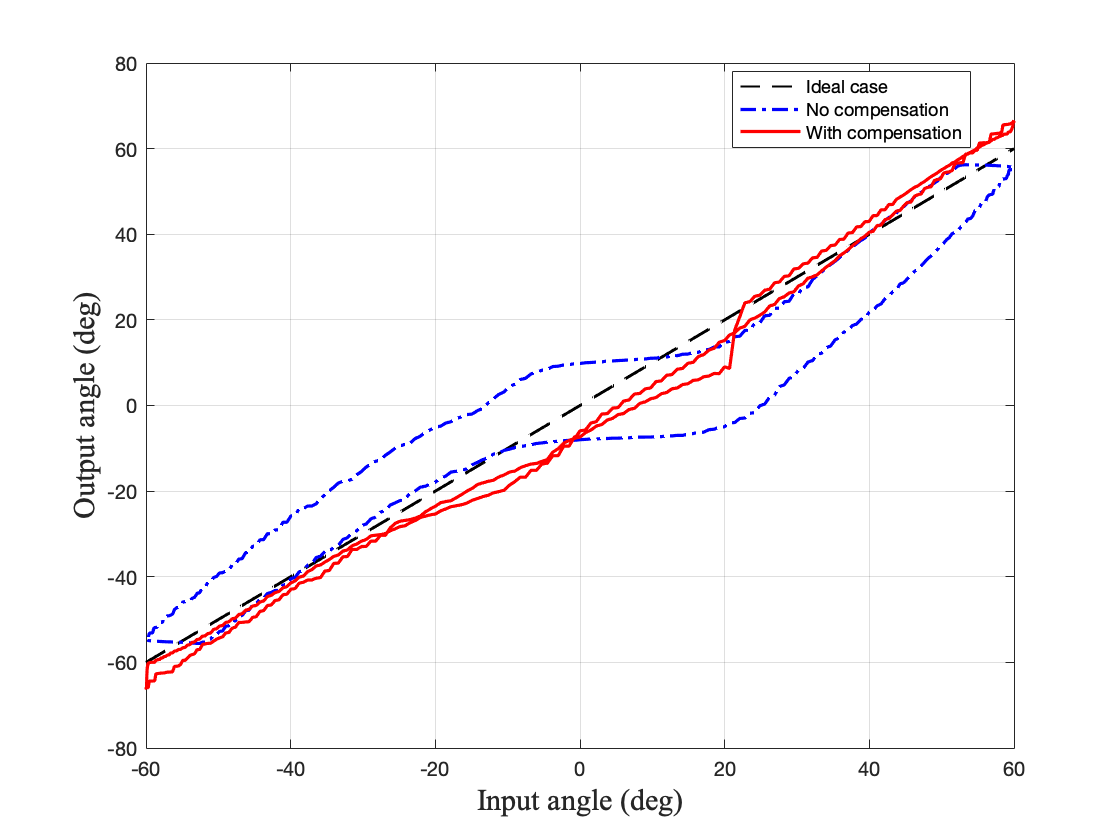}\label{fig:LR4}}
			\hspace{-15pt}
			\subfigure[$\phi_1$ (or $\phi_2$) = $60^\circ$]{\includegraphics[scale=0.1]{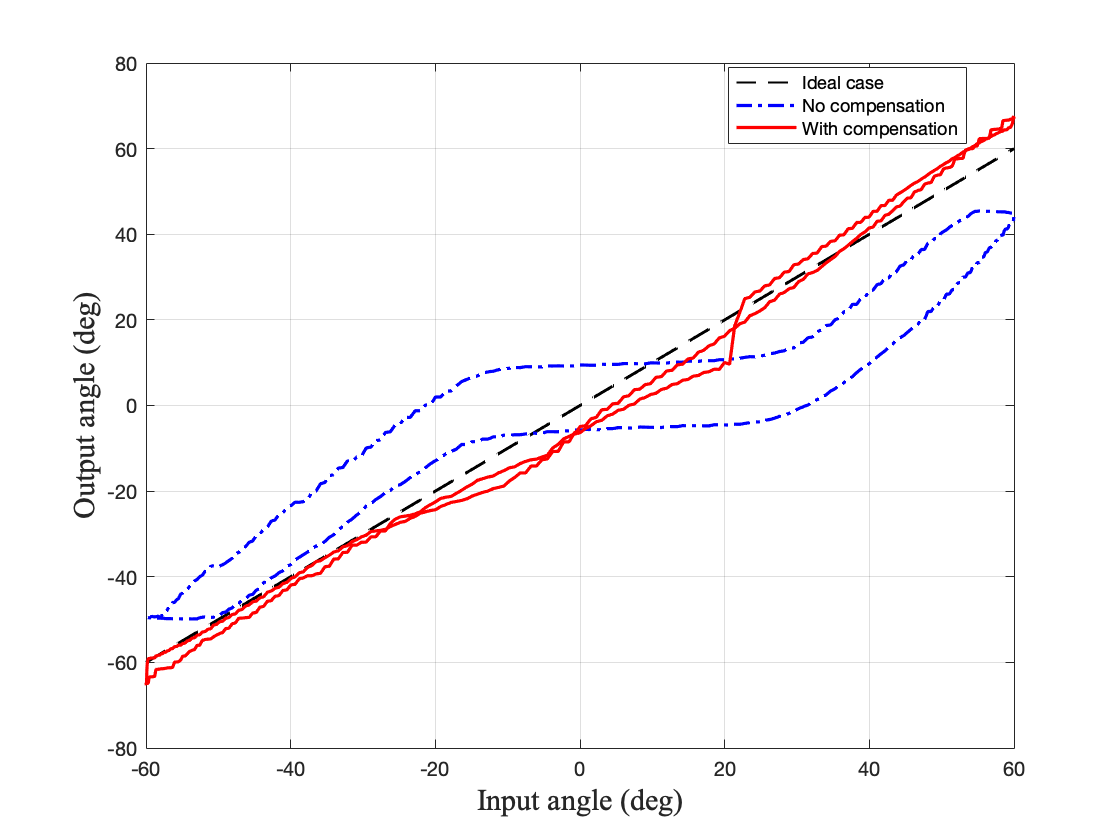}\label{fig:LR5}}
			
			\caption{The result of the non-linear behavior compensation for one catheter: The black dot is the ground truth. The blue dot is without compensation. The red dot is with our compensation. The first row shows time versus output angle, and the second row shows the input angle versus the output angle for the anterior-posterior knob $\phi_1$ motion, thus the left-right knob $\phi_2$ is fixed as addressed.  \label{fig:result_1dof}
			}
		\end{center}
		
	\end{figure*}

	\yh{Figure\,\ref{fig:result_1dof} shows an exemplary result for 1-DOF by {\it a periodic motion}; The first row showed the time versus output angle, which displayed the desired input as the black dotted line), {\it no compensation} is the blue dotted line, and {\it with compensation} is the red line.
		Further, the input angle versus the output angle is in the second row. In the ideal case the line would be linear, representing matching input and output. The compensated output (a red-dotted line) was closely located in the linear line. The overall performance evaluation is described in Table\,\ref{table:1d_fixed}. 
		Our proposed method improved PTPE by 62 to 66 $\%$ and RMSE by 71 to 76 $\%$. 
		The 1-DOF by {\it non-periodic motion} was not shown in this figure due to limited page, but the overall results are explained in Table\,\ref{table:1d_all}. The 1-DOF results for {\it A periodic motion} shows PTPE improved by $63\,\%$ and RMSE improved by $73\,\%$, while {\it Non-periodic motion} shows PTPE improved by $48\,\%$ and RMSE improved by $63\,\%$.}

	
	
	
	
\begin{figure*}
		\subfigcapskip = -8pt
		\begin{center}
			
			\hspace{-15pt}
			\subfigure[Time  versus  output  angle,  $\phi_1$]{\includegraphics[scale=0.12]{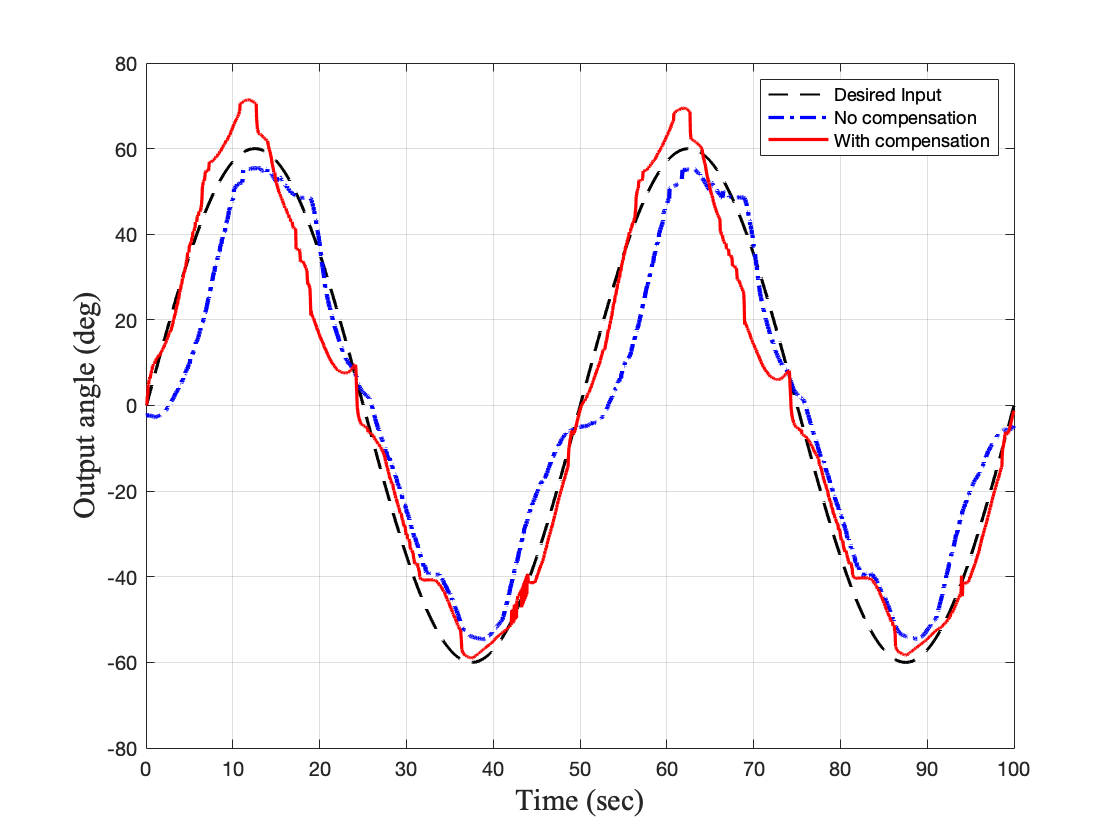}\label{fig:1AP}}
			\hspace{-15pt}
			\subfigure[Time  versus  output  angle,  $\phi_2$]{\includegraphics[scale=0.12]{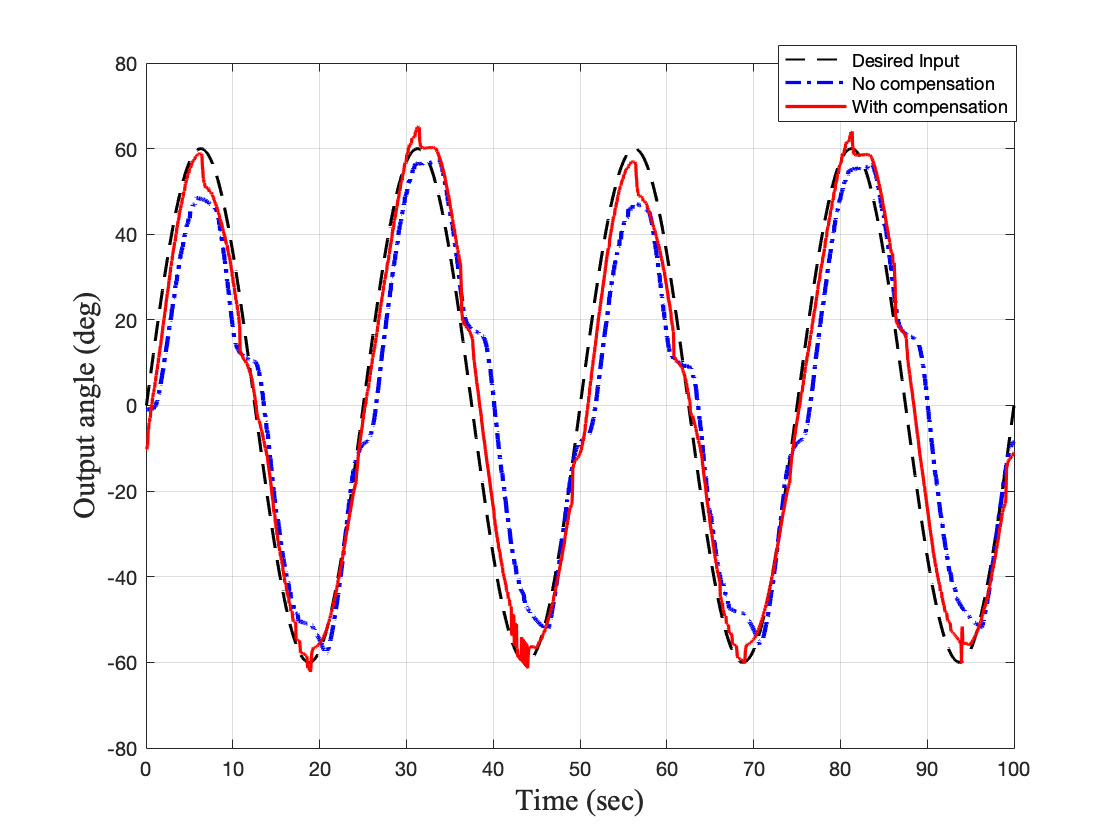}\label{fig:1LR}}
			\hspace{-15pt}
			\subfigure[Time  versus  output  angle, $\phi_1$]
			{\includegraphics[scale=0.12]{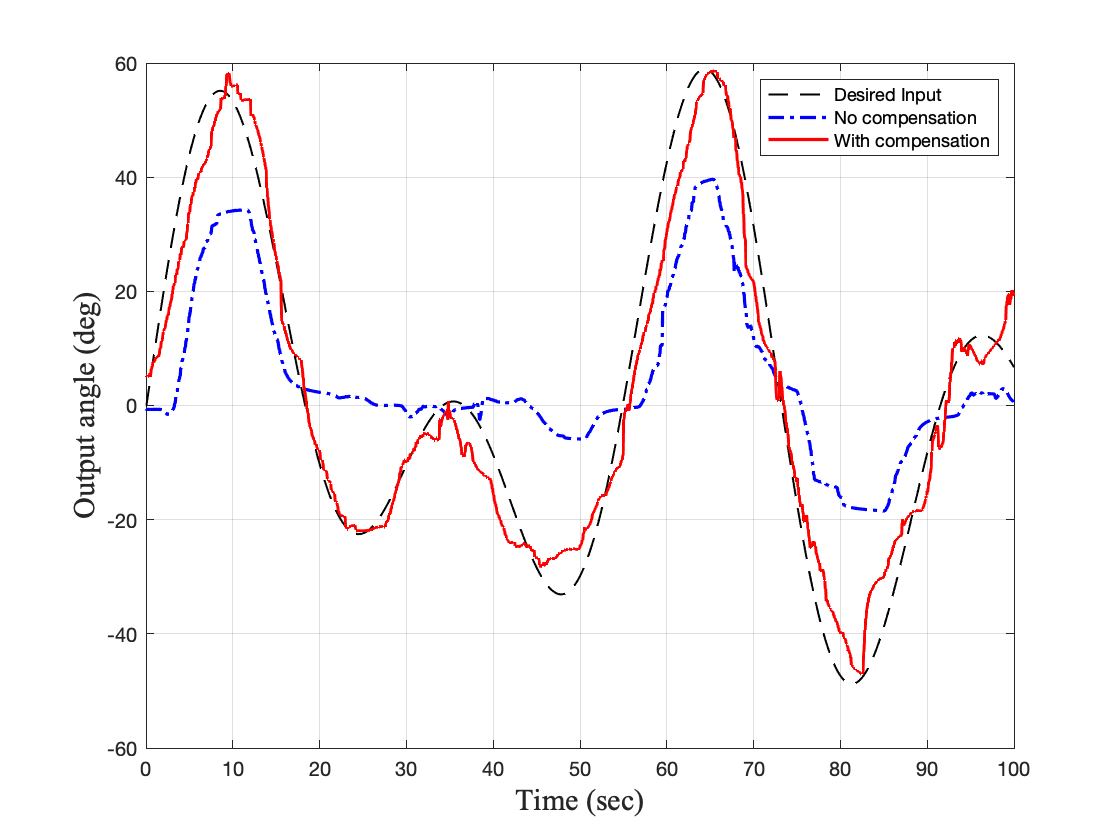}}
			\hspace{-15pt}
			\subfigure[Time  versus  output  angle, $\phi_2$]
			{\includegraphics[scale=0.12]{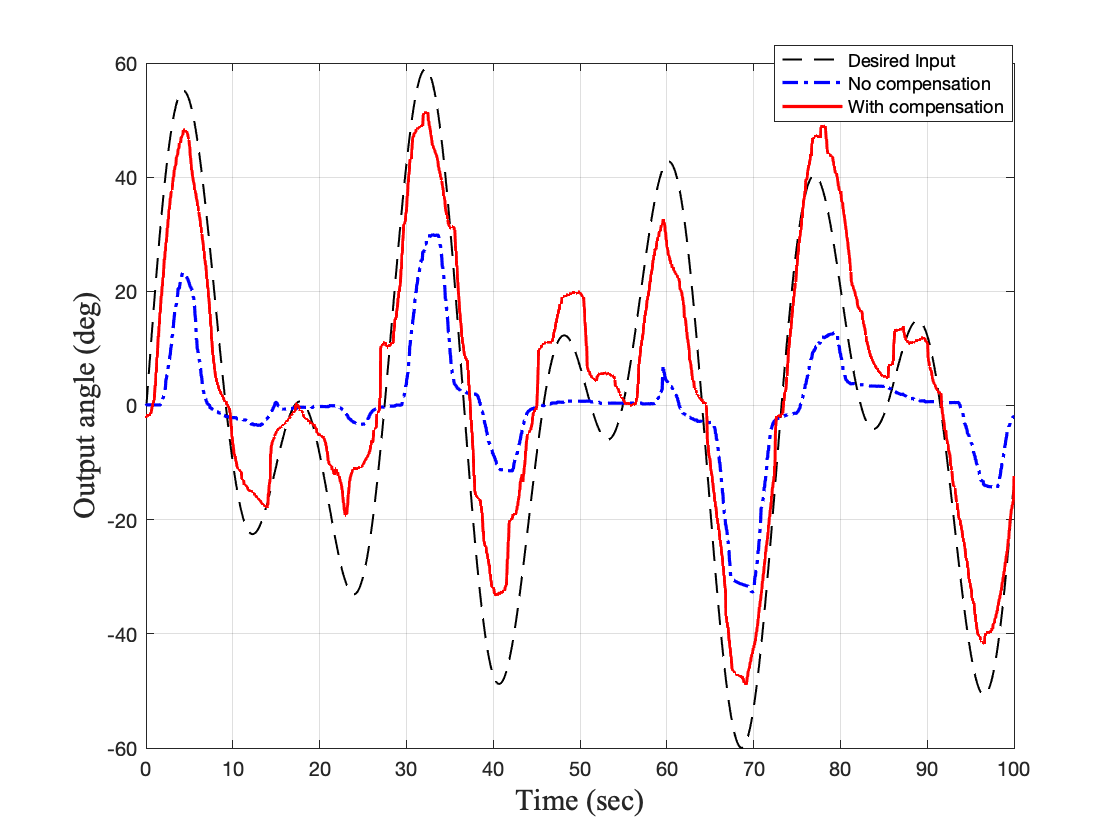}}
			
			
			\hspace{-15pt}
			\subfigure[Input  versus  output angle,  $\phi_1$]{\includegraphics[scale=0.12]{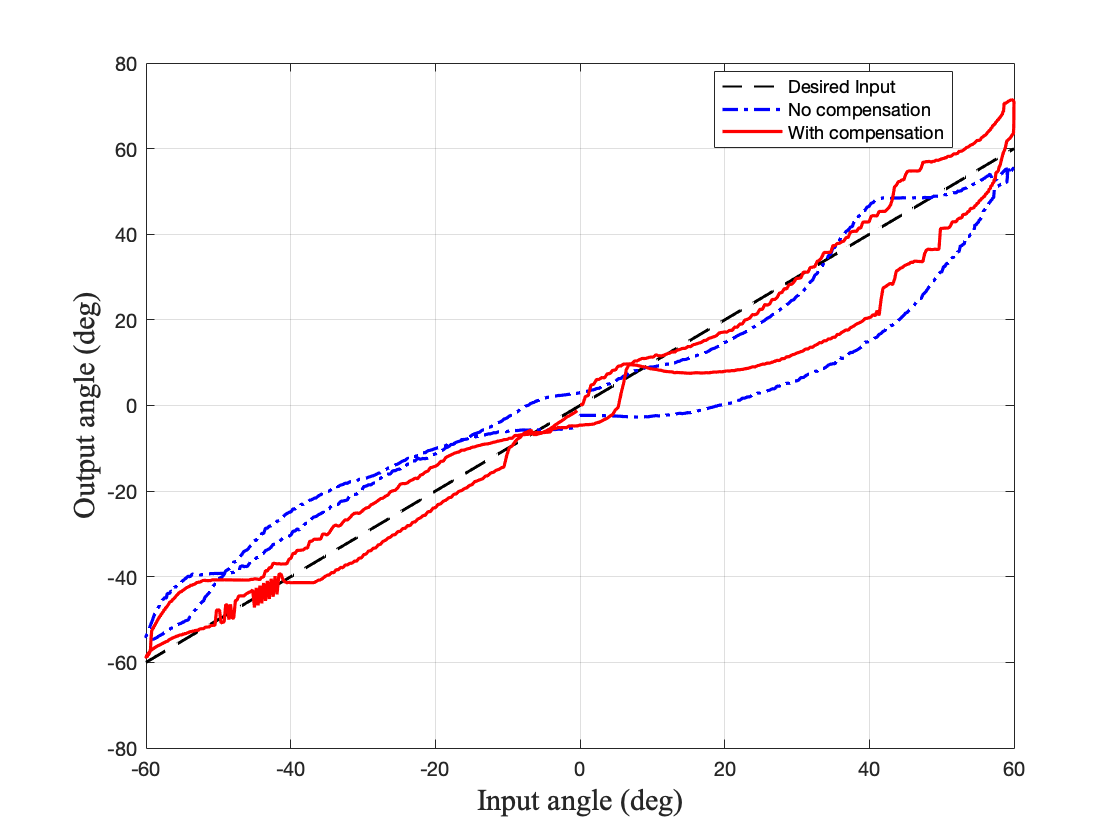}}
			\hspace{-15pt}
			\subfigure[Input  versus  output angle,  $\phi_2$]{\includegraphics[scale=0.12]{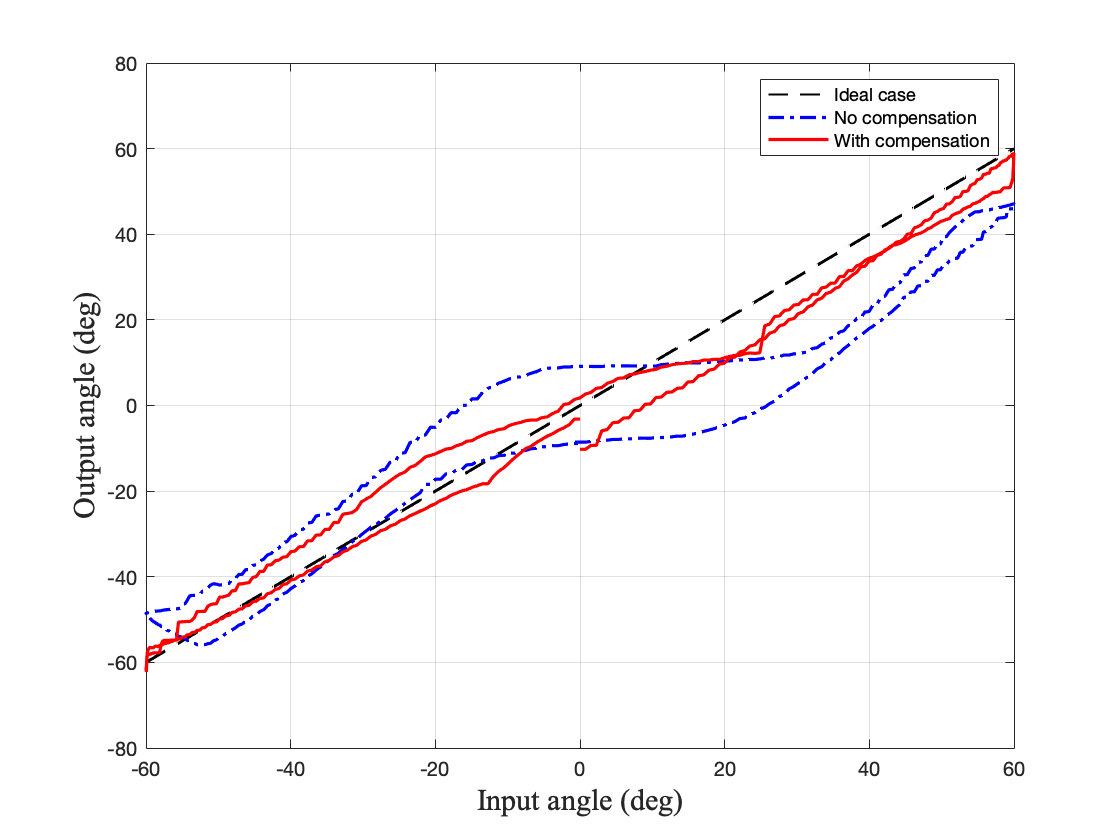}}
			\hspace{-15pt}
			\subfigure[ Input  versus  output angle, $\phi_1$]{\includegraphics[scale=0.12]{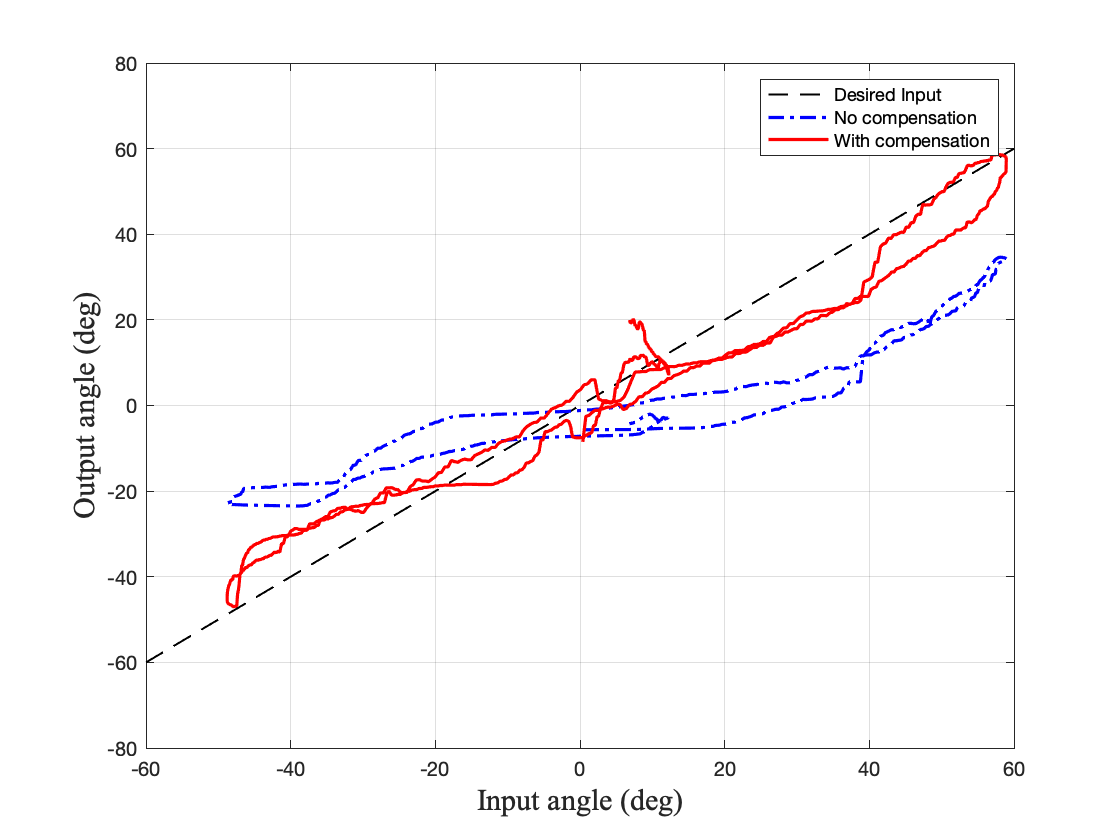}}
			\hspace{-15pt}
			\subfigure[ Input  versus  output angle, $\phi_2$]{\includegraphics[scale=0.12]{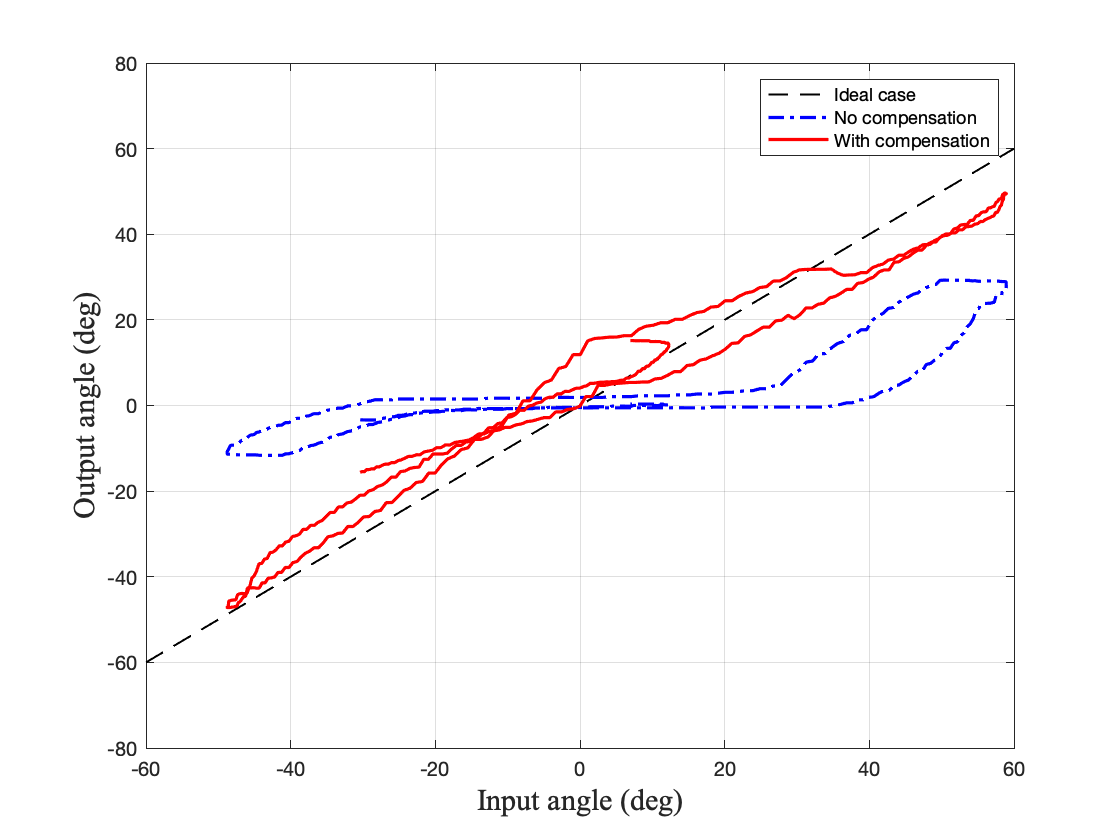}}
			\caption{ This demonstrates 2~DOF results for one catheter: The black dot is the ground truth. The blue dot is without compensation. The red dot is with our compensation. The first two column is for a periodic input. The last two column is for a non-periodic input. The first row shows time versus output angle, $\phi_1$ and $\phi_2$ have different frequency as the results demonstrated. The second row shows the input angle versus the output angle for $\phi_1$ and $\phi_2$ motions.
				\label{fig:result_2dof_total}}
		\end{center}
	\end{figure*}

	\yh{Figure\,\ref{fig:result_2dof_total} demonstrated 2-DOF results for one of three catheters. The first row shows time versus output angle for $\phi_1$ and $\phi_2$ having different frequency. The second row shows the input angle versus the output angle for $\phi_1$ and $\phi_2$ motions. The first two column shows {\it a periodic} input motion, and the the last two column shows {\it a non-periodic} input motion. PTPE and RMSE results are shown in Table\,\ref{table:2d_all}; PTPE and RMSE are improved by $40\%$ and $58\%$ for {\it A periodic} 2D-inputs, while PTPE and RMSE are improved by $44\%$ and $55\%$ for {\it A non-perioric} 2D-inputs, respectively.}
	

	\begin{table*}
		\centering
		\caption{PTPE and RMSE for 1-DOF while another DOF fixed: Periodic}\label{table:1d_fixed}
		\scalebox{0.9}{\begin{tabular}{|c||c|c||c|c||c|c|}
				\hline               Fixed angle from other DOF & \multicolumn{2}{|c||}{$0^\circ$} &\multicolumn{2}{|c||}{$\pm 30^\circ$}
				&\multicolumn{2}{|c|}{$\pm 60^\circ$}\\
				\hline
				Metrics             & Peak-to-peak    & RMSE  & Peak-to-peak    & RMSE  & Peak-to-peak    & RMSE 
				\\ \hline
				No compensation (deg)   &    (66.0, 7.0)             &  
				(21.2, 2.5)       &    
				(65.2, 9.7)             &  
				(20.9, 3.0)       &
				(68.9, 8.2)             &  
				(23.9, 3.5)       \\ \hline
				With compensation (deg) &   (21.9, 5.4)             & 
				(6.1, 2.0)     &    
				(21.9, 5.9)             &  
				(5.3, 2.3)     &    
				(23.9, 3.5)             &   
				(5.7, 0.8)     \\ \hline
				Improvement rate ($\%$) &   
				66.82            &   
				71.21            &   
				66.47            &  
				74.65            &   
				62.51            &  
				76.27   \\   
				\hline
		\end{tabular}}
	\end{table*}

	\begin{table}
		\centering
		\caption{PTPE and RMSE for 1-DOF: periodic and non-periodic inputs}\label{table:1d_all}
		\scalebox{0.82}{\begin{tabular}{|c||c|c||c|c|}
				\hline                Test inputs & \multicolumn{2}{|c||}{Periodic 1D-Inputs} &\multicolumn{2}{|c|}{Non-periodic 1D-Inputs}\\
				\hline
				Metrics             & Peak-to-peak    & RMSE  & Peak-to-peak    & RMSE  
				\\ \hline
				No compensation (deg)   &    (66.4, 7.2)             &  (22.2, 2.9)       &    (57.9, 1.5)             &  (17.2, 0.5)       \\ \hline
				With compensation (deg) &   (24.0, 5.2)             &   (5.9, 1.7)     &    (30.0, 4.6)             &   (6.3, 1.1)     \\ \hline
				Improvement rate ($\%$) &   63.8             &   73.2   &   48.1             &   63.8   \\   
				\hline
		\end{tabular}}
	\end{table}

	\begin{table}
		\centering
		\caption{PTPE and RMSE for 2-DOFs: periodic and non-periodic inputs}\label{table:2d_all}
		\scalebox{0.84}{\begin{tabular}{|c||c|c||c|c|}
				\hline                Test inputs & \multicolumn{2}{|c||}{Periodic 2D-Inputs} &\multicolumn{2}{|c|}{Non-periodic 2D-Inputs}\\
				\hline
				Metrics             & Peak-to-peak    & RMSE  & Peak-to-peak    & RMSE  
				\\ \hline
				No compensation   &    (64.3, 5.7)             &  (23.8, 3.5)       &    (65.7, 5.3)             &  (18.9, 1.7)       \\ \hline
				With compensation &   (38.3, 7.2)             &   (10.0, 2.6)     &   (37.0, 7.5)             &   (8.42, 1.4)     \\ \hline
				Improvement rate ($\%$) &   40.5             &   58.0   &   43.7             &   55.3   \\   
				\hline
		\end{tabular}}
	\end{table}

	\section{Discussion}
	
	The experimental results showed that the proposed method is effective to compensate for non-linear hysteresis. Figure \ref{fig:result_1dof} shows  typical examples; there is a delay in the dead zone before compensation, but it is reduced after compensation. In addition, the desired input angle $60$ degree was not reached before compensation, but it can reach $60$ degree. Lastly, the shape of the graph shows a {straightened} line shape similar to the ideal case after compensation, which is shown in Figure\,\ref{fig:result_1dof} from (f) to (j).
	
	
	Our model is based on the piecewise linear approximation, so there exists a limitation, which 
	{can present as} a jerk motion in transitions. For example, our approach generate the shape changes when entering or exiting the dead zone. However, there is a discrepancy between real phenomena in Figure\,\ref{fig:merged_relationship} and our model in Figure\,\ref{fig:approximated_simple_model} based on experimental results. Also the backlash hysteresis section is not straight line in real phenomena. For this reason, the jerk movement can be observed. 
	
	The result of two DOF also showed good improvement, however, it was not as good as one DOF test. It seems that there exist coupling effects between two DOFs in mechanical structures, which is not easily separable in practice. Most commercial products of TSM have commonly complicated mechanical structures inside due to considerations of multiple uses ({\em e.g.,} ultrasound image, grasping tools). Our previous work\,\citet{kim2020automatic} demonstrated the method to compensate the plastic torsion effects due to coupling using attached external sensors. However, this paper proposed a simple calibration method using the motor current only, which is limited to detect coupling effects.

	
	
	
	Our experimental setup has the shape constraints, which is the shape of the sheath in a straight line. First, this paper focused on understanding the relationship between motor current and hysteresis so that model parameters can be identified without external sensors in practice.
	It is evident that the sheath shape change is another challenging problem. However, the shape changes mostly affect the width of dead zone $D$ based on our observation. Since the dead zone $D$ can be identified by motor current, if motions can be operated in target environments, our proposed method might be able to update $D$ values according to shape changes.

	\section{Conclusion}	 
	
	This paper proposed a simplified piecewise linear model to compensate non-linear hysteresis of both dead zone and backlash in multiple tendon-sheath-mechanism. Moreover, a simple parameter identification method was introduced for practical settings ({\em e.g.,} surgical room) based on our systematic validation of relationship between hysteresis curve and motor current. 
	Through the relationship between the non-linear hysteresis and the behavior of the motor current, the range of the dead zone ${\bf D}$ and the size of the backlash hysteresis ${\bf B}$ were obtained, and the slope of the straight line $\omega$ was obtained through mechanical properties or a data-driven method. Accordingly, the height of the dead zone ${\bf H}$ can be followed. Based on our proposed method, all parameters were easily calibrated for the TSM robotic manipulators without external sensors, which are not always possible in clinical environments.
	\yh{Lastly, our methods were validated with three different ICE catheters, attached to robotic manipulators. The 1-DOF and 2-DOF controls with periodic/non-periodic input motions were demonstrated, and both the peak-to-peak and the mean squared errors were significantly reduced.}



	
	\section*{Disclaimer}
	The concepts and information presented in this abstract/paper are based on research results that are not commercially available. Future availability cannot be guaranteed.

	\newpage
	{
		\small
		\bibliography{references_icra}
	}

\end{document}